\pdfoutput=1

\documentclass[11pt]{article}

\usepackage{EACL2023}

\usepackage{times}
\usepackage{latexsym}

\usepackage[T1]{fontenc}

\usepackage[utf8]{inputenc}

\usepackage{microtype}

\usepackage{inconsolata}

\usepackage{amsmath}
\usepackage{float}
\usepackage{multirow}
\usepackage{xspace}
\usepackage{graphicx}
\usepackage{pifont}
\usepackage{booktabs}
\usepackage{arydshln}

\newcommand{\Model}{MAPL\xspace}
\newcommand{\Modelblind}{MAPL-blind\xspace}
\newcommand{\maple}{\includegraphics[height=.75\baselineskip]{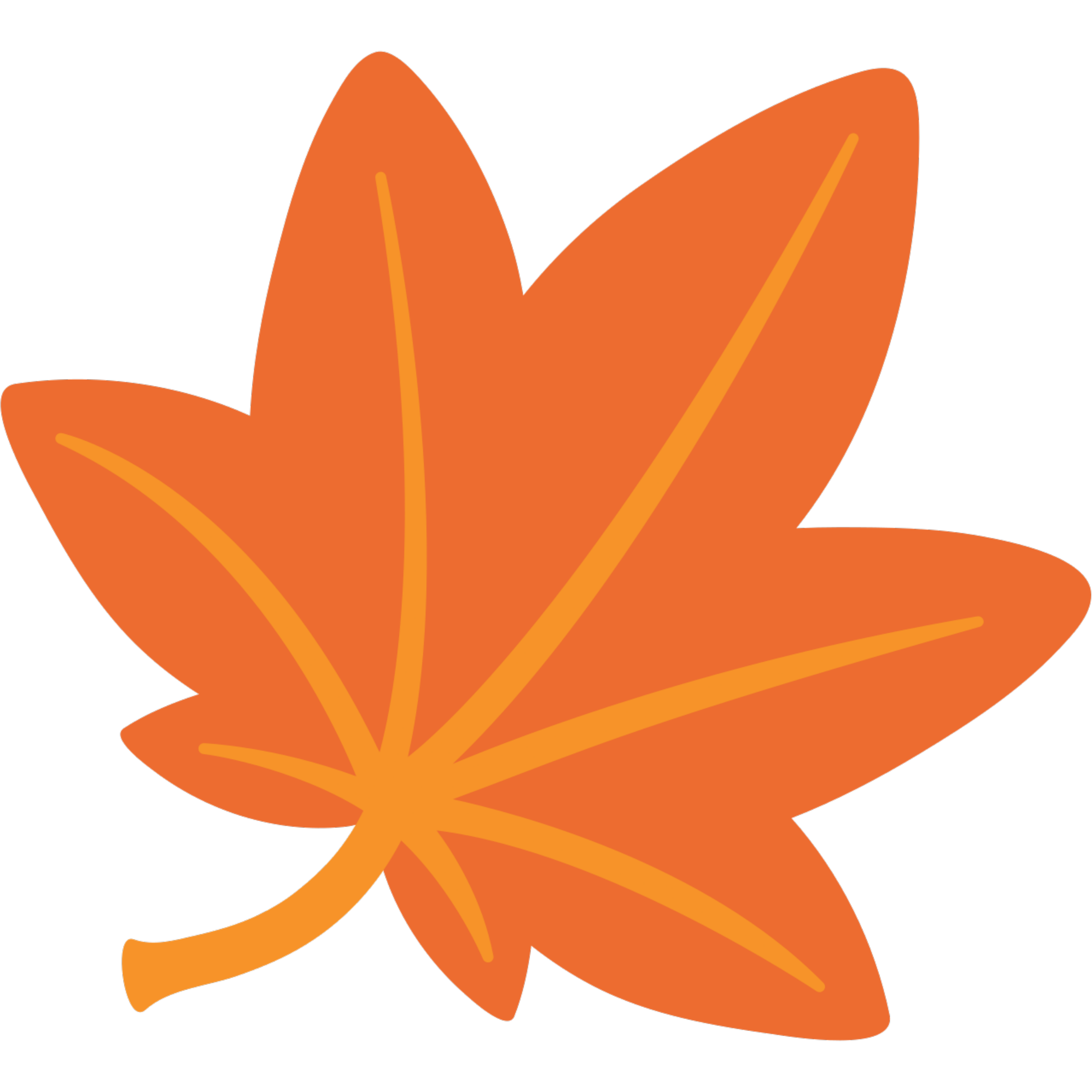}\xspace}
\newcommand{\Frozen}{Frozen$^\ast$\xspace}
\newcommand{\PICa}{PICa$^\ast$\xspace}

\title{\Model{}\maple{}: Parameter-Efficient Adaptation of Unimodal Pre-Trained Models for Vision-Language Few-Shot Prompting}

\newcommand{\mila}{\triangle}
\newcommand{\snow}{\diamondsuit}
\newcommand{\dm}{\heartsuit}
\newcommand{\sait}{\clubsuit}
\newcommand{\cifar}{\dagger}

\author{Oscar Mañas$^{\mila}$~ Pau Rodriguez$^{*,\snow}$~ Saba Ahmadi$^{*,\mila}$ \\
\textbf{Aida Nematzadeh$^{\dm}$~ Yash Goyal$^{\sait}$~ Aishwarya Agrawal$^{\mila,\cifar}$} \\[4pt]
$^{\mila}$Mila, Université de Montréal~\;$^{\snow}$ServiceNow Research~\;$^{\dm}$DeepMind \\
~$^{\sait}$Samsung - SAIT AI Lab, Montreal~\;$\cifar$Canada CIFAR AI Chair \\[4pt]
\texttt{oscar.manas@mila.quebec} \\}

\begin{document}

\maketitle
\def\thefootnote{*}\footnotetext{denotes equal contribution.}\def\thefootnote{\arabic{footnote}}

\begin{abstract}

Large pre-trained models have proved to be remarkable zero- and (prompt-based) few-shot learners in unimodal vision and language tasks. We propose \Model, a simple and parameter-efficient method that reuses frozen pre-trained unimodal models and leverages their strong generalization capabilities in multimodal vision-language (VL) settings. \Model learns a lightweight mapping between the representation spaces of unimodal models using aligned image-text data, and can generalize to unseen VL tasks from just a few in-context examples. The small number of trainable parameters makes \Model effective at low-data and in-domain learning. Moreover, \Model's modularity enables easy extension to other pre-trained models. Extensive experiments on several visual question answering and image captioning benchmarks show that \Model achieves superior or competitive performance compared to similar methods while training orders of magnitude fewer parameters. \Model can be trained in just a few hours using modest computational resources and public datasets. We release our code and pre-trained model weights at \url{https://github.com/mair-lab/mapl}.

\end{abstract}

\section{Introduction}
Over the past few years, natural language processing and computer vision have witnessed impressive progress in learning models capable of transferring to unseen tasks or benchmarks~\citep{brown2020language,zhang2022opt,radford2021learning,jia2021scaling}. Recently referred to as foundation models~\citep{bommasani2021opportunities}, these can be adapted to a wide range of \emph{unimodal} vision and language tasks without any additional training.

In this work, we study reusing such powerful \emph{unimodal} foundation models for \emph{multimodal} vision-language (VL) downstream tasks. In particular, we propose to connect a vision encoder, such as CLIP~\citep{radford2021learning}, to an autoregressive language model (LM), such as GPT~\cite{radford2018improving,radford2019language,brown2020language}, with minimal additional training on multimodal data. Our goal is to obtain a single VL model that can leverage the in-context learning abilities~\citep{brown2020language} of the pre-trained LM to generalize to unseen VL tasks from just a few examples. 

One challenge in connecting vision encoders with LMs is aligning the visual and textual representation spaces. Recent works have approached this by adapting the LM to visual representations, either by fine-tuning the entire LM~\citep{dai2022enabling} or training adapter layers~\citep{eichenberg2021magma,alayrac2022flamingo}. These systems are computationally expensive to train as they have a large number of learnable parameters (hundreds of millions to a few billions) and use large-scale multimodal training data. On the other hand, Frozen~\citep{tsimpoukelli2021multimodal} keeps the LM frozen, thus learning ${\sim}10\times$ less parameters than the above methods. However, it requires training a visual encoder from scratch, which is also computationally expensive.

Differently, we aim to reuse large pre-trained unimodal models while keeping them completely frozen and free of adapter layers. We present \textbf{\Model} (\textbf{M}ultimodal \textbf{A}daptation of \textbf{P}re-trained vision and \textbf{L}anguage models), a simple and parameter-efficient VL model capable of tackling unseen VL tasks. \Model learns a lightweight mapping between the representation spaces of pre-trained unimodal models. \Model has orders of magnitude fewer parameters than previous methods (including Frozen) and can be trained in just a few hours. Moreover, \Model's modularity makes it general-purpose and easily extensible to newer and/or better pre-trained models. We evaluate \Model on various image captioning and visual question answering (VQA) benchmarks and compare with Frozen~\citep{tsimpoukelli2021multimodal} in a controlled setup. \Model significantly outperforms Frozen and achieves competitive performance compared to other methods~\citep{eichenberg2021magma,dai2022enabling} trained on comparably sized data.

We further investigate the parameter efficiency of \Model by training on only 1\% of multimodal data (thousands of examples); we call this setting \emph{low-data} learning. We also study \emph{in-domain} learning: training on image-text pairs from the same domain as the downstream task domains. We train \Model directly on 100\% and 1\% of in-domain data for each downstream task, without first pre-training on large-scale domain-agnostic data. Thus, we train specialized versions of \Model for each downstream domain. Such low-data and in-domain learning are particularly useful when it is difficult to pre-train on large-scale domain-agnostic data. We found \Model to be more effective than Frozen trained under the same settings.

To summarize, our contributions are: 1) we introduce \Model, a parameter-efficient method capable of tackling unseen VL tasks, which can be trained using only modest computational resources and public datasets; 2) we conduct extensive experiments spanning various image captioning and VQA benchmarks, demonstrating \Model achieves superior or competitive performance compared to similar methods while training orders of magnitude fewer parameters; and 3) we further investigate the parameter-efficiency of \Model in two settings: low-data and in-domain. Our experiments show that \Model is more effective than the considered methods in both settings.

\section{Related Work}
\paragraph{Fine-tuning based VL methods.} 
A popular family of VL methods are based on the pre-training + fine-tuning paradigm. These methods are either encoder-only~\cite{lu2019vilbert,tan2019lxmert,chen2019uniter,li2020oscar,zhang2021vinvl} or encoder-decoder methods~\citep{cho2021unifying,wang2021simvlm,jin2021good,li2021align} and use transformer-based architectures. These transformers are first pre-trained on domain-agnostic image-text pairs (e.g., Conceptual Captions~\citep{sharma2018conceptual}) using self-supervised objectives, and then fine-tuned for each downstream task (e.g., VQA, image captioning). More recent models that are designed specifically for the task of image captioning use large pre-trained LMs (e.g., GPT-2~\citep{radford2019language}) and fine-tune these models with image-caption pairs~\citep{chen2021visualgpt,mokady2021clipcap,DBLP:journals/corr/abs-2201-12723}. While all these approaches yield state-of-the-art performance for the tasks they are fine-tuned on, the learned model weights are highly specialized for a single task and cannot transfer to new tasks with zero or few examples. Differently, \Model reuses the same set of weights for all downstream tasks without any additional training.

\vspace{-7pt}
\paragraph{Few-shot learning based VL methods.}
Most similar to \Model are methods that tackle unseen VL tasks in a zero/few-shot manner, by leveraging the in-context learning abilities of large pre-trained LMs (e.g., GPT-3~\citep{brown2020language}. These methods connect a vision encoder with a pre-trained LM to tackle VL tasks. Some methods~\cite{dai2022enabling,hao2022language} achieve this connection by fine-tuning the entire LM on image-text data, while others only train adapter layers inserted into the LM~\citep{eichenberg2021magma}. The vision encoder is pre-trained and kept frozen in both cases. Concurrent work Flamingo~\cite{alayrac2022flamingo} pushes this idea even further by scaling up the amount of training data and the LM size. While inserting adapter layers requires training fewer parameters compared to fine-tuning the entire LM, the number of trainable parameters is still >100M; in contrast, \Model only has 3.4M trainable parameters. Additionally, inserting adapter layers is not straightforward since it requires modifying the computational graph of the LM; \Model only adds an external mapping network, which is easier to incorporate on top of pre-trained models. On the other hand, Frozen~\cite{tsimpoukelli2021multimodal} keeps the pre-trained LM frozen and instead trains a vision encoder from scratch.
This approach does not scale well with larger vision encoders (Sec.~\ref{sec:ablations}). \Model keeps both the vision encoder and the LM frozen (thus further reducing the number of trainable parameters) and only learns a lightweight mapping network to connect both frozen models.
Similar to \Model, concurrent work LiMBeR~\cite{merullo2022linearly} also proposes to connect a frozen vision encoder with a frozen LM but using a linear mapping, which is not as parameter- and compute-efficient as \Model (Sec.~\ref{sec:ablations}).

\begin{figure*}[ht]
    \centering
    \includegraphics[width=\linewidth]{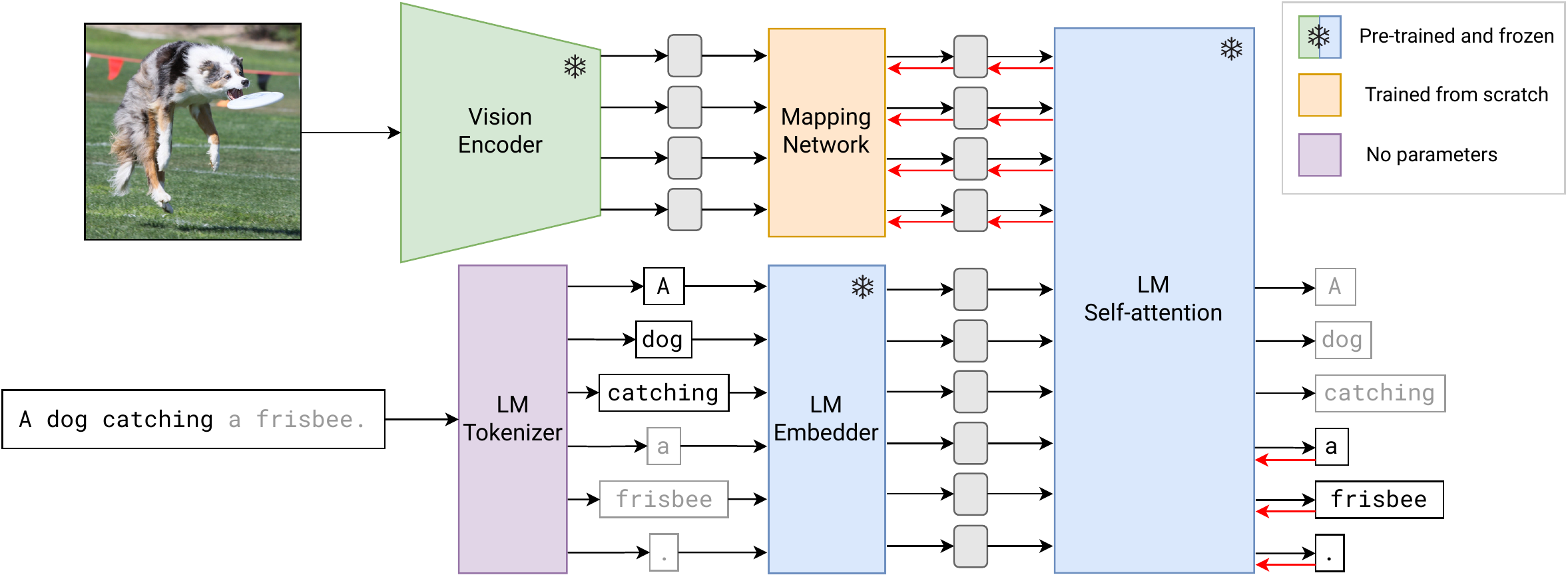}
    \caption{\Model leverages a pre-trained vision encoder and a pre-trained LM, and learns a small \textit{mapping network} to convert visual features into token embeddings. During training, only the mapping network is updated, keeping the vision encoder and the LM frozen (red arrows indicate gradient flow). At inference time, the system can take as input an arbitrary sequence of interleaved images and text, and generates free-form text as output.}
    \vspace{-12pt}
    \label{fig:method_overview}
\end{figure*}

\vspace{-7pt}
\paragraph{Mapping networks.}
\Model trains a mapping network to align the visual and textual representations of the visual encoder and the LM, respectively. The architecture of our mapping network has some similarities with that in ClipCap~\cite{mokady2021clipcap} and the Perceiver Resampler in Flamingo~\cite{alayrac2022flamingo}. They all share a core transformer stack and a fixed number of learned constant embeddings. However, \Model's mapping network is specifically designed to be parameter-efficient while maintaining expressivity (Sec.~\ref{sec:architecture}), containing only 3.4M parameters -- orders of magnitude fewer than ClipCap's (43M) and Flamingo's (194M).

\section{Method}
\Model is a vision-language (VL) multimodal model capable of generating text from a combination of visual and textual inputs. Our model builds on top of pre-trained vision-only and language-only models and leverages their strong generalization capabilities (e.g., zero-shot transfer, in-context learning) to tackle unseen VL tasks. \Model is agnostic to the choice of these pre-trained unimodal models as long as they show such capabilities (Sec.~\ref{sec:ablations}). Concretely, \Model maps the image representations from a vision encoder's output embedding space to a LM's token embedding space, so that the LM can be conditioned both on visual and textual information. To this end, we train a \textit{mapping network} with an image captioning objective (Sec.~\ref{sec:architecture}, \ref{sec:training}), while keeping the weights of the vision encoder and the LM frozen. Once the mapping network is trained, \Model can be prompted with a few examples of unseen VL tasks and predict the response via text generation (Sec.~\ref{sec:fewshot_transfer}). The overall model architecture is depicted in Figure~\ref{fig:method_overview}.

\subsection{Architecture}
\label{sec:architecture}

\noindent \textbf{Pre-trained vision encoder.}
The vision encoder extracts a compact representation from an image. We use a CLIP~\citep{radford2021learning} pre-trained vision encoder, which is trained on web-scale data and has shown strong zero-shot transfer capabilities to unseen image domains. In particular, we use CLIP's ViT-L/14 backbone~\citep{dosovitskiy2020image} since we empirically found it yields the best downstream VL performance among all variants. We use the flattened grid of spatial features ($16 \times 16$) before the final projection layer and the representation corresponding to the \texttt{[class]} token, resulting in a sequence of $L_i = 257$ vectors of dimensionality $D_i = 1024$ each. This sequence of vectors is then fed to the mapping network.

\vspace{-3pt}
\paragraph{Pre-trained autoregressive language model.}
Given an input text, the language model (LM) predicts its most likely completion by generating free-form text.
For our LM, we use a pre-trained GPT-J model~\cite{gpt-j}~\footnote{We also experiment with an OPT model, see Sec.~\ref{sec:ablations}.}, a publicly-released 6B-parameter autoregressive LM trained on the Pile dataset~\cite{gao2020pile}. We chose this LM due to its strong in-context learning abilities, similar to that of GPT-3~\citep{brown2020language} (which is not publicly available). The LM takes as input a text string, which is first divided into a sequence of discrete tokens by the LM's tokenizer. Each token is then individually transformed into a continuous embedding (of size $D_o = 4096$) by the LM's embedder. The sequence of token embeddings is fed to the self-attention layers in the LM's transformer block (using causal attention), which outputs a sequence of categorical distributions over the token vocabulary. Finally, a decoding mechanism generates free-form text from these distributions (greedy decoding in our case).

\begin{figure}[t]
    \centering
    \includegraphics[width=\linewidth]{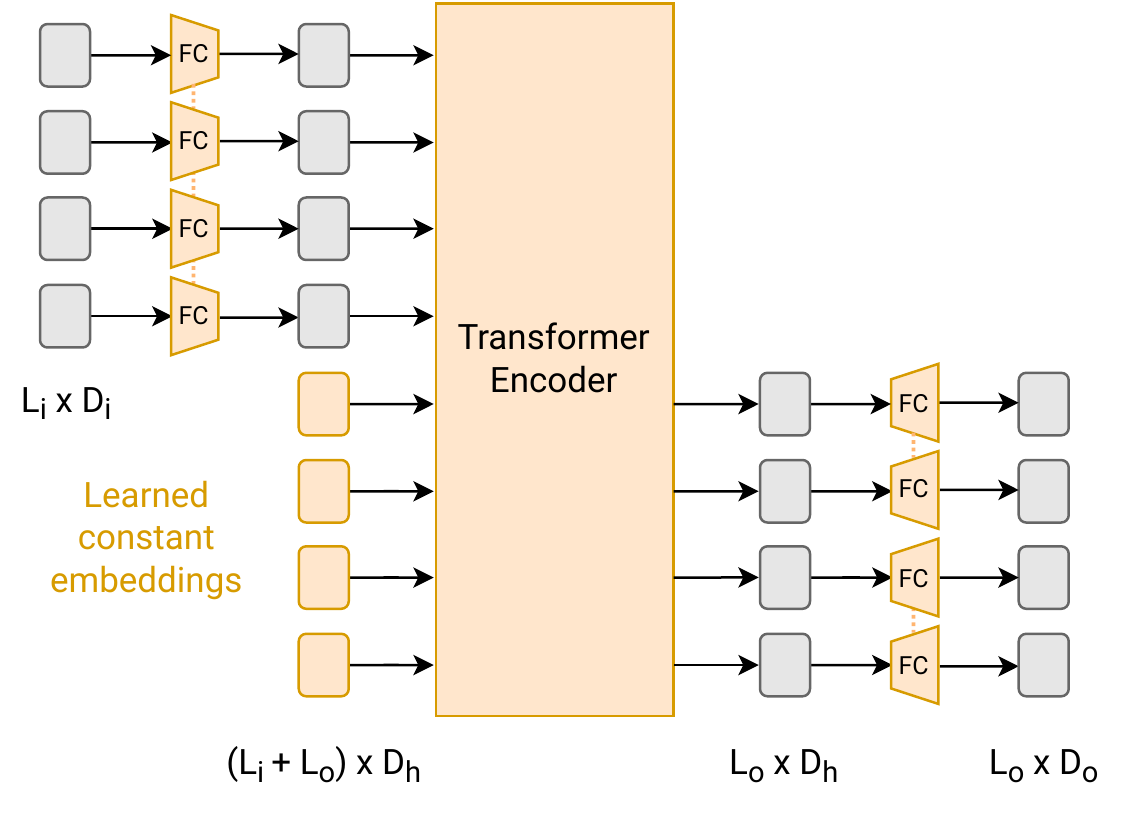}
    \vspace{-20pt}
    \caption{The mapping network takes a flattened grid of $L_i$ visual features of dimension $D_i$ each 
    from the vision encoder and transforms it into a sequence of token embeddings of length $L_o$ and dimension $D_o$, where $D_o$ is the token embedding dimension of the LM. Note that the parameters are shared across fully-connected (FC) layers, on both sides of the encoder transformer.}
    \vspace{-10pt}
    \label{fig:mapping_network}
\end{figure}

\vspace{-3pt}
\paragraph{Mapping network.}
The mapping network transforms a sequence of visual features from the vision encoder to a sequence of continuous embeddings which can be consumed by the LM's transformer. We design our mapping network considering the trade-off between expressivity (to learn a good mapping) and parameter count. Our architecture is based on a transformer encoder with 4 layers and 8 heads each. This transformer could directly take a sequence of projected visual features (from $D_i$ to $D_o$) and output a sequence of embeddings of size $D_o$. However, in order to keep a low parameter count, we decouple the transformer hidden size $D_h$ from the visual feature size $D_i$ and the LM embedding size $D_o$ by introducing a dimensionality bottleneck (Figure~\ref{fig:mapping_network}). In particular, each visual feature is first linearly projected from $D_i = 1024$ to $D_h = 256$ using a set of fully-connected (FC) layers. This sequence of projected features is then fed to the transformer, and the output representations are linearly projected from $D_h$ to $D_o = 4096$ using another set of FC layers. To further reduce the parameter count of our mapping network, we share parameters across all FC layers in each set.

Yet another idea we use in our mapping network is to decouple the output sequence length of the transformer ($L_o$) from the input sequence length ($L_i$). We do this to obtain a much smaller $L_o = 32$ compared to $L_i = 257$, in order to reduce the computational complexity in the subsequent LM's self-attention layers, which in turn speeds up training and inference time. To achieve this decoupling, inspired by DETR~\citep{carion2020end}, we concatenate a small and fixed number ($L_o$) of learned constant embeddings with the input sequence of the transformer and only use the output representations corresponding to these constant embeddings (Figure~\ref{fig:mapping_network}). Note that these output representations are conditioned on the input visual features via cross-attention in the transformer. The resulting mapping network architecture is shown in Figure~\ref{fig:mapping_network}. In total, our mapping network contains only 3.4M parameters. Since this is the only trainable component of our model, \Model has orders of magnitude fewer total trainable parameters than existing methods such as Frozen (40.3M) or Flamingo (10.2B).

\subsection{Training}
\label{sec:training}

Following previous works \cite{tsimpoukelli2021multimodal,eichenberg2021magma}, we train our model using a standard language modeling objective on image captions with teacher forcing~\cite{lamb2016professor}, i.e., we minimize the negative log-likelihood of the reference captions under the LM conditioned on the corresponding images. We only train the mapping network (from scratch) while keeping the vision encoder and the LM entirely frozen. This preserves the pre-trained models' capabilities while making the system modular and parameter-efficient. Even though the LM's weights are kept frozen, gradients are still back-propagated through its self-attention layers to train the mapping network.

\subsection{Zero- and Few-shot Evaluation}
\label{sec:fewshot_transfer}

Once the mapping network is trained, \Model can tackle unseen VL tasks by prompting the LM with a combination of visual and textual inputs. We study zero-shot transfer to unseen image captioning benchmarks and few-shot transfer (via in-context learning) to the unseen task of visual question answering (VQA). For image captioning, we simply feed the mapped image embedding to the LM and start generating a caption. For zero-shot VQA, following \citet{tsimpoukelli2021multimodal}, we feed the mapped image embedding followed by the text \texttt{\color{teal} ``Please answer the question. Question: \{question\} Answer:''}\footnote{Here \texttt{\{question\}} indicates a placeholder which gets replaced by the corresponding question in each example. Same applies to \texttt{\{answer\}} in the few-shot setting.} and start generating the answer. For $n$-shot VQA, we select $n$ support examples $(image, question, answer)$ from the training set at random, and prepend them to the query; for each support example, we concatenate the mapped image embedding with the text \texttt{\color{teal} ``Please answer the question. Question: \{question\} Answer: \{answer\}''}.

\section{Experiments}
\subsection{Experimental settings}
\label{sec:experimental_setting}

\paragraph{Evaluation benchmarks.} We evaluate \Model on several VL benchmarks spanning VQA and image captioning. Note that our model is never trained for the task of VQA. For VQA, we evaluate on the validation splits of VQAv2~\cite{goyal2017making}, OK-VQA~\cite{marino2019ok}, TextVQA~\cite{singh2019towards} and VizWiz-VQA~\cite{gurari2018vizwiz}, and report performance using VQA accuracy (after the standard normalization~\cite{antol2015vqa}). For image captioning, we evaluate on the Karpathy-test split~\cite{karpathy2015deep} of COCO Captions~\cite{chen2015microsoft}, and the validation splits of Conceptual Captions (CC)~\cite{sharma2018conceptual}\footnote{Due to broken image URLs, we only managed to download 13K out of 15K validation images.}, TextCaps~\cite{sidorov2020textcaps} and VizWiz-Captions~\cite{gurari2018vizwiz}, and report performance using the BLEU@4, ROUGE-L, METEOR, CIDEr and SPICE metrics.

\vspace{-5pt}
\paragraph{Training settings.} We consider two settings to train our mapping network: domain-agnostic and in-domain training (described below). For each of these settings, we also study low-data learning by training our model on randomly sampled subsets of 1\% training image-text pairs. Such low-data learning is useful when it is difficult to train models on large-scale data due to constraints on compute resources, data availability, etc. 

For \textbf{domain-agnostic training}, we use the CC dataset, which is gathered by automatically scraping images and their corresponding alt-text fields from web pages. Thus, this dataset is not as clean as manually-curated datasets such as COCO Captions (e.g., the caption may not describe the image). Nevertheless, due to its large size (3.3M) and great diversity, it is the most commonly used dataset for domain-agnostic pre-training of VL models. However, for our model -- having orders of magnitude less trainable parameters than other methods --, we observed the negative effect of noise in CC to be stronger than the positive effect of its large size (Sec.~\ref{sec:noisy_data}). Therefore, we train \Model on a filtered version of CC (CC-clean) consisting of the top 398K most similar image-text pairs ranked by CLIP's image-text similarity score.\footnote{We selected a threshold on CLIP's similarity score such that the size of the filtered dataset is comparable to the size of manually curated datasets such as COCO Captions.} For completeness, we also report \Model's performance when trained on the unfiltered CC dataset.

For \textbf{in-domain training}, we use image-caption pairs that come from the same domain as the downstream task domains, i.e., they have similar image and language distributions as those in the downstream datasets. For the image captioning downstream task, this amounts to the IID setting. The in-domain image captioning and VQA dataset pairs we consider are shown in Table~\ref{tab:indomain_datasets}. Each pair uses the same set of images, and focuses on the same set of VL skills; for instance, scene understanding (COCO Caps and VQAv2), reading and reasoning about text in images (TextCaps and TextVQA), understanding images captured by visually-impaired users (VizWiz-Caps and VizWiz-VQA), thus leading to similar image and language distributions across image-captioning and VQA. We train \Model on both 100\% and 1\% of in-domain image-caption data and evaluate on all downstream benchmarks (including out-of-domain ones, e.g., VizWiz-VQA when trained on COCO Caps). Such in-domain training can be useful when it is difficult to first train on large-scale domain-agnostic data and then adapt to in-domain data by either fine-tuning or few-shot prompting.

\begin{table}[ht]
    \centering
    \resizebox{0.48\textwidth}{!}{%
    \begin{tabular}{l|cccc}
        \toprule
         & \textbf{VQAv2} & \textbf{OK-VQA} & \textbf{TextVQA} & \textbf{VizWiz-VQA} \\
        \midrule
        \textbf{COCO Caps} & \ding{51} & \ding{51} & & \\
        \textbf{TextCaps} & & & \ding{51} & \\
        \textbf{VizWiz-Caps} & & & & \ding{51} \\
        \bottomrule
    \end{tabular}%
    }
    \vspace{-5pt}
    \caption{In-domain dataset pairs.}
    \vspace{-10pt}
    \label{tab:indomain_datasets}
\end{table}

\begin{table*}[t]
    \resizebox{\textwidth}{!}{%
    \begin{tabular}{l|c|c||ccc|ccc|ccc|ccc|ccc}
        \toprule
         & \textbf{Trainable} & \textbf{Training} & \multicolumn{3}{c|}{\textbf{n-shot VQAv2}} & \multicolumn{3}{c|}{\textbf{n-shot OK-VQA}} & \multicolumn{3}{c|}{\textbf{n-shot TextVQA}} & \multicolumn{3}{c|}{\textbf{n-shot VizWiz-VQA}} & \multicolumn{3}{c}{\textbf{n-shot Overall}} \\
         & \textbf{params} & \textbf{examples} & \textbf{0} & \textbf{4} & \textbf{8} & \textbf{0} & \textbf{4} & \textbf{8} & \textbf{0} & \textbf{4} & \textbf{8} & \textbf{0} & \textbf{4} & \textbf{8} & \textbf{0} & \textbf{4} & \textbf{8} \\
        \hline
        \multicolumn{1}{l}{} & \multicolumn{1}{c}{} & \multicolumn{1}{c}{} & \multicolumn{15}{c}{\textbf{Existing methods using domain-agnostic training}} \\
        Frozen & 40.3M$^\dagger$ & 3.3M & 29.50 & 38.20 & - & 5.90 & 12.60 & - & - & - & - & - & - & - & - & - & - \\
        MAGMA\xspace$_\text{CC12M}$ & 243M$^\dagger$ & 3.8M & 36.90 & 45.40 & - & 13.90 & 23.40 & - & - & - & - & 5.60 & 10.60 & - & - & - & - \\
        VLKD\xspace$_\text{CC3M}$ & 406M & 3.3M & 38.60 & - & - & 10.50 & - & - & - & - & - & - & - & - & - & - & - \\
        LiMBeR-CLIP$^\ddagger$ & 12.6M$^\dagger$ & 3.3M & 33.33 & 40.34 & - & - & - & - & - & - & - & - & - & - & - & - & - \\
        \hdashline
        Flamingo$^\ddagger$ & 10.2B & >2.1B & - & - & - & 50.60 & 57.40 & 57.50 & 35.00 & 36.50 & 37.30 & - & - & - & - & - & - \\
        \hline
        \multicolumn{1}{l}{} & \multicolumn{1}{c}{} & \multicolumn{1}{c}{} & \multicolumn{15}{c}{\textbf{100\% domain-agnostic training}} \\
        \Modelblind$_\text{CC-clean}$ & 3.4M & 374K & 20.62 & 35.01 & 35.11 & 4.84 & 14.68 & 14.28 & 3.68 & 5.43 & 5.82 & 3.18 & 8.65 & 9.55 & 8.08 & 15.94 & 16.19 \\
        \Frozen$_\text{CC-clean}$ & 40.3M & 374K & 25.98 & 37.80 & 38.52 & 5.51 & 18.86 & 19.91 & 5.11 & 6.15 & 6.30 & 4.33 & 11.28 & 16.68 & 10.23 & 18.52 & 20.35 \\
        \Model$_\text{CC-clean}$ & 3.4M & 374K & \textbf{33.54} & \textbf{45.13} & \textbf{45.21} & \textbf{13.84} & \textbf{24.25} & \textbf{23.93} & \textbf{8.26} & \textbf{8.88} & \textbf{8.77} & \textbf{11.72} & \textbf{18.46} & \textbf{19.52} & \textbf{16.84} & \textbf{24.18} & \textbf{24.36} \\
        \hline
        \multicolumn{1}{l}{} & \multicolumn{1}{c}{} & \multicolumn{1}{c}{} & \multicolumn{15}{c}{\textbf{1\% domain-agnostic training}} \\
        \Frozen$_\text{CC-clean}$ & 40.3M & 3.7K & 26.22 & 36.69 & 37.41 & 5.50 & \textbf{18.76} & \textbf{20.51} & 5.71 & \textbf{7.19} & 7.53 & 3.83 & \textbf{11.71} & \textbf{16.66} & 10.31 & \textbf{18.58} & \textbf{20.53} \\
        \Model$_\text{CC-clean}$ & 3.4M & 3.7K & \textbf{30.80} & \textbf{37.38} & \textbf{37.95} & \textbf{8.77} & 18.18 & 19.15 & \textbf{6.40} & 7.07 & \textbf{7.74} & \textbf{5.68} & 9.26 & 10.58 & \textbf{12.91} & 17.97 & 18.85 \\
        \hline
        \multicolumn{1}{l}{} & \multicolumn{1}{c}{} & \multicolumn{1}{c}{} & \multicolumn{15}{c}{\textbf{100\% in-domain training}} \\
        \PICa & 0 & 0 & 20.61 & 46.86 & 47.80 & 11.84 & \textbf{31.28} & \textbf{33.07} & - & - & - & - & - & - & - & - & - \\
        \hdashline
        \Frozen$_\text{COCO}$ & 40.3M & 414K & 32.09 & 38.90 & 39.42 & 9.81 & 20.72 & 21.83 & 7.54 & 6.82 & 6.74 & 5.87 & 12.07 & 17.35 & 13.82 & 19.63 & 21.33 \\
        \Frozen$_\text{TextCaps}$ & 40.3M & 103K & 32.49 & 37.39 & 38.03 & 11.34 & 19.87 & 20.82 & 8.83 & 7.33 & 7.51 & 6.25 & 12.26 & 16.86 & 14.73 & 19.21 & 20.80 \\
        \Frozen$_\text{VizWiz}$ & 40.3M & 110K & 26.93 & 37.38 & 37.91 & 5.85 & 19.12 & 20.64 & 6.38 & 7.44 & 7.47 & 5.57 & 13.06 & 18.06 & 11.18 & 19.25 & 21.02 \\
        \hdashline
        \Model$_\text{COCO}$ & 3.4M & 414K & \textbf{43.51} & \textbf{48.75} & \textbf{48.44} & \textbf{18.27} & 31.13 & 31.63 & 10.99 & 11.10 & 11.08 & \textbf{14.05} & 17.72 & 19.18 & 21.70 & \textbf{27.17} & \textbf{27.58} \\
        \Model$_\text{TextCaps}$ & 3.4M & 103K & 38.83 & 43.34 & 43.43 & 16.33 & 25.07 & 25.92 & \textbf{22.27} & \textbf{19.53} & \textbf{19.75} & 12.31 & 16.69 & 18.18 & \textbf{22.43} & 26.15 & 26.82 \\
        \Model$_\text{VizWiz}$ & 3.4M & 110K & 32.80 & 42.94 & 43.20 & 11.70 & 24.91 & 25.73 & 9.27 & 10.36 & 10.23 & 10.42 & \textbf{20.63} & \textbf{23.10} & 16.05 & 24.71 & 25.56 \\
        \hline
        \multicolumn{1}{l}{} & \multicolumn{1}{c}{} & \multicolumn{1}{c}{} & \multicolumn{15}{c}{\textbf{1\% in-domain training}} \\
        \Frozen$_\text{COCO}$ & 40.3M & 4.1K & 30.18 & 37.23 & 37.89 & 9.33 & 19.60 & 20.71 & 7.43 & 7.65 & 7.67 & 4.37 & 12.00 & 16.48 & 12.83 & 19.12 & 20.69 \\
        \Frozen$_\text{TextCaps}$ & 40.3M & 1.0K & 32.09 & 36.72 & 37.25 & 10.75 & 18.85 & 19.51 & 8.17 & 7.57 & 7.28 & 5.39 & 11.79 & 16.20 & 14.10 & 18.73 & 20.06 \\
        \Frozen$_\text{VizWiz}$ & 40.3M & 1.1K & 29.62 & 37.30 & 37.87 & 7.57 & 19.36 & 20.60 & 7.16 & 7.17 & 7.25 & 4.53 & \textbf{12.51} & \textbf{17.56} & 12.22 & 19.08 & \textbf{20.82} \\
        \hdashline
        \Model$_\text{COCO}$ & 3.4M & 4.1K & \textbf{37.69} & \textbf{40.42} & \textbf{40.84} & \textbf{13.92} & \textbf{21.66} & \textbf{22.41} & 8.30 & 6.96 & 6.84 & \textbf{6.94} & 10.72 & 12.43 & \textbf{16.71} & \textbf{19.94} & 20.63 \\
        \Model$_\text{TextCaps}$ & 3.4M & 1.0K & 33.57 & 36.70 & 36.87 & 12.46 & 17.45 & 18.21 & \textbf{9.34} & \textbf{8.29} & \textbf{8.62} & 6.54 & 9.58 & 11.62 & 15.48 & 18.00 & 18.83 \\
        \Model$_\text{VizWiz}$ & 3.4M & 1.1K & 31.88 & 36.81 & 37.04 & 9.59 & 17.64 & 17.64 & 7.25 & 5.99 & 6.04 & 4.73 & 9.48 & 11.33 & 13.36 & 17.48 & 18.01 \\
        \bottomrule
    \end{tabular}%
    }
    \vspace{-7pt}
    \caption{Evaluation on few-shot VQA. For MAGMA\xspace$_\text{CC12M}$ and VLKD\xspace$_\text{CC3M}$, we report their best results when training only on domain-agnostic data (CC12M and CC3M, respectively). ($^\dagger$) indicates our informed estimation. ($^\ddagger$) indicates concurrent work.}
    \vspace{-15pt}
    \label{tab:vqa}
\end{table*}

\vspace{-5pt}
\paragraph{Training details.} For Conceptual Captions, TextCaps and VizWiz-Captions, we carve out a minival split consisting of 6\% of training examples and train on the remaining 94\%; for COCO Captions, we use the Karpathy-val split as minival. We use the AdamW optimizer with $\beta_1 = 0.9$, $\beta_2 = 0.95$, and a weight decay of $0.01$. The learning rate is increased linearly from 0 to $3\times10^{-4}$ ($7\times10^{-4}$ for OPT-based models) over the first $1500$ steps ($15$ for 1\% of data) and kept constant for the rest of training. We use a batch size of $128$ and we do early stopping based on the minival loss. We do not add any special tokens at the beginning of sentence, as GPT-J was not trained with \texttt{<BOS>} tokens. In order to fit a 6B-parameter LM into GPU memory, we use DeepSpeed ZeRO~\cite{rajbhandari2020zero} stage 2 optimizations. Freezing the LM's weights also brings massive savings in GPU memory during training, as fine-tuning with an Adam-based optimizer would require at least $4\times$ GPU memory to store gradients, average, and squared average of the gradients. The whole system was trained on 4 A100 (40GB) GPUs for about 4 hours (for the CC-clean dataset). Unless otherwise stated, we repeat the experiments with two different random seeds and report the average performance.

\vspace{-5pt}
\paragraph{Existing methods and baselines.}
We report the performance of several baselines and existing methods. First, to verify that the LM in \Model is not ignoring the visual input, inspired by \citet{tsimpoukelli2021multimodal}, we train a blind version of \Model (\Modelblind) where the input images are replaced with zeros but the mapping network weights are still trained (to serve as prompt-tuning for the LM). Second, to estimate the upper-bound on how well we can do in VQA by representing images with text (rather than with continuous embeddings), we evaluate PICa~\cite{yang2021empirical}, which directly prompts the LM with image captions, followed by questions for VQA. We reimplement PICa (denoted \PICa) using \Model's LM (and evaluate on VQAv2 and OK-VQA using ground-truth COCO captions) for controlled comparison. Third, we compare \Model with Frozen~\citep{tsimpoukelli2021multimodal}, as this is the most similar method to ours that also uses a frozen LM. We reimplement Frozen (denoted \Frozen) using \Model's LM for controlled comparison. Lastly, we report the performance of other methods similar to \Model: MAGMA~\cite{eichenberg2021magma}, VLKD~\cite{dai2022enabling}, LiMBeR~\cite{merullo2022linearly}, ClipCap~\cite{mokady2021clipcap} and the published numbers from Frozen~\cite{tsimpoukelli2021multimodal}.\footnote{We only add results which are reported on the same dataset splits as in \Model.} Note that all these methods (unless otherwise noted) are trained on domain-agnostic data, so we only compare with \Model trained on CC-clean. For completeness, we also report results from Flamingo~\cite{alayrac2022flamingo}, which has orders of magnitude more learnable parameters than \Model and is trained on considerably more data.\looseness=-1

\begin{table*}[t]
    \resizebox{\textwidth}{!}{\begin{tabular}{l|c|c||cc|cc|cc|cc|cc}
        \toprule
         & \textbf{Trainable} & \textbf{Training} & \multicolumn{2}{c|}{\textbf{CC}} & \multicolumn{2}{c|}{\textbf{COCO}} & \multicolumn{2}{c|}{\textbf{TextCaps}} & \multicolumn{2}{c|}{\textbf{VizWiz-Caps}} & \multicolumn{2}{c}{\textbf{Overall}} \\
         & \textbf{params} & \textbf{examples} & \textbf{B@4} & \textbf{CIDEr} & \textbf{B@4} & \textbf{CIDEr} & \textbf{B@4} & \textbf{CIDEr} & \textbf{B@4} & \textbf{CIDEr} & \textbf{B@4} & \textbf{CIDEr} \\
        \hline
        \multicolumn{1}{l}{} & \multicolumn{1}{c}{} & \multicolumn{1}{c}{} & \multicolumn{10}{c}{\textbf{Existing methods using domain-agnostic training}} \\
        ClipCap\xspace$_\text{CC3M}$ & 43M & 3.3M & - & 71.82 & - & - & - & - & - & - & - & - \\
        VLKD\xspace$_\text{CC3M}$ & 406M & 3.3M & - & - & 18.20 & 61.10 & - & - & - & - & - & - \\
        \hline
        \multicolumn{1}{l}{} & \multicolumn{1}{c}{} & \multicolumn{1}{c}{} & \multicolumn{10}{c}{\textbf{100\% domain-agnostic training}} \\
        \Modelblind$_\text{CC-clean}$ & 3.4M & 374K & 0.35 & 5.05 & 2.75 & 5.75 & 1.35 & 2.15 & 1.50 & 1.80 & 1.49 & 3.69 \\
        \Frozen$_\text{CC-clean}$ & 40.3M & 374K & 2.45 & 22.60 & 5.25 & 13.90 & 2.65 & 4.60 & 2.05 & 2.65 & 3.10 & 10.94 \\
        \Model$_\text{CC-clean}$ & 3.4M & 374K & \textbf{6.75} & \textbf{79.75} & \textbf{12.30} & \textbf{54.30} & \textbf{5.80} & \textbf{22.95} & \textbf{4.95} & \textbf{20.95} & \textbf{7.45} & \textbf{44.49} \\
        \hline
        \multicolumn{1}{l}{} & \multicolumn{1}{c}{} & \multicolumn{1}{c}{} & \multicolumn{10}{c}{\textbf{1\% domain-agnostic training}} \\
        \Frozen$_\text{CC-clean}$ & 40.3M & 3.7K & 0.75 & 6.55 & 3.05 & 5.25 & 1.70 & 1.65 & 1.50 & 1.40 & 1.75 & 3.71 \\
        \Model$_\text{CC-clean}$ & 3.4M & 3.7K & \textbf{1.75} & \textbf{19.65} & \textbf{5.80} & \textbf{17.85} & \textbf{2.70} & \textbf{5.40} & \textbf{2.15} & \textbf{4.85} & \textbf{3.10} & \textbf{11.94} \\
        \hline
        \multicolumn{1}{l}{} & \multicolumn{1}{c}{} & \multicolumn{1}{c}{} & \multicolumn{10}{c}{\textbf{100\% in-domain training}} \\
        \Frozen$_\text{COCO}$ & 40.3M & 414K & 0.65 & 9.05 & 20.05 & 61.35 & 6.95 & 11.75 & 5.45 & 6.20 & 8.28 & 22.09 \\
        \Frozen$_\text{TextCaps}$ & 40.3M & 103K & 0.20 & 3.55 & 4.05 & 6.70 & 8.85 & 16.95 & 4.40 & 5.25 & 4.38 & 8.11 \\
        \Frozen$_\text{VizWiz}$ & 40.3M & 110K & 0.25 & 4.40 & 3.75 & 6.05 & 4.10 & 5.65 & 19.00 & 76.85 & 6.78 & 23.24 \\
        ClipCap\xspace$_\text{COCO}$ & 43M & 414K & - & - & 33.53 & 113.08 & - & - & - & - & - & - \\
        \hdashline
        \Model$_\text{COCO}$ & 3.4M & 414K & \textbf{2.25} & \textbf{34.50} & \textbf{36.45} & \textbf{125.20} & 16.60 & 41.40 & 18.00 & 41.35 & \textbf{18.33} & \textbf{60.61} \\
        \Model$_\text{TextCaps}$ & 3.4M & 103K & 0.90 & 13.05 & 9.80 & 28.65 & \textbf{18.35} & \textbf{62.55} & 11.20 & 31.85 & 10.06 & 34.03 \\
        \Model$_\text{VizWiz}$ & 3.4M & 110K & 0.90 & 18.80 & 13.55 & 48.35 & 11.35 & 31.20 & \textbf{34.70} & \textbf{141.30} & 15.13 & 59.91 \\
        \hline
        \multicolumn{1}{l}{} & \multicolumn{1}{c}{} & \multicolumn{1}{c}{} & \multicolumn{10}{c}{\textbf{1\% in-domain training}} \\
        \Frozen$_\text{COCO}$ & 40.3M & 4.1K & 0.25 & 3.60 & 6.20 & 12.80 & 2.80 & 3.15 & 2.85 & 2.30 & 3.03 & 5.46 \\
        \Frozen$_\text{TextCaps}$ & 40.3M & 1.0K & 0.10 & 2.60 & 1.65 & 2.80 & 3.65 & 5.00 & 2.00 & 2.25 & 1.85 & 3.16 \\
        \Frozen$_\text{VizWiz}$ & 40.3M & 1.1K & 0.20 & 3.40 & 2.90 & 3.20 & 3.35 & 3.45 & 12.70 & 40.55 & 4.79 & 12.65 \\
        \hdashline
        \Model$_\text{COCO}$ & 3.4M & 4.1K & \textbf{0.80} & \textbf{12.10} & \textbf{19.65} & \textbf{65.90} & 7.00 & 12.85 & 6.20 & 9.60 & \textbf{8.41} & \textbf{25.11} \\
        \Model$_\text{TextCaps}$ & 3.4M & 1.0K & 0.30 & 3.90 & 4.10 & 8.05 & \textbf{8.35} & \textbf{16.90} & 5.00 & 7.25 & 4.44 & 9.03 \\
        \Model$_\text{VizWiz}$ & 3.4M & 1.1K & 0.20 & 3.90 & 2.95 & 4.80 & 3.45 & 5.05 & \textbf{18.40} & \textbf{71.10} & 6.25 & 21.21 \\
        \bottomrule
    \end{tabular}%
    }
    \vspace{-7pt}
    \caption{Evaluation on image captioning. For VLKD\xspace$_\text{CC3M}$, we report their best results when training only on domain-agnostic data (CC3M).}
    \vspace{-5pt}
    \label{tab:caption}
\end{table*}
\begin{table*}[t]
    \resizebox{\textwidth}{!}{%
    \begin{tabular}{l|c||cccccccc|cc}
        \toprule
         & \textbf{Training} & \textbf{VQAv2} & \textbf{OK-VQA} & \textbf{TextVQA} & \textbf{VizWiz-VQA} & \textbf{CC} & \textbf{COCO} & \textbf{TextCaps} & \textbf{VizWiz-Caps} & \multicolumn{2}{c}{\textbf{Overall}} \\
         & \textbf{examples} & \textbf{4-shot} & \textbf{4-shot} & \textbf{4-shot} & \textbf{4-shot} & \textbf{CIDEr} & \textbf{CIDEr} & \textbf{CIDEr} & \textbf{CIDEr} & \textbf{4-shot} & \textbf{CIDEr} \\
        \midrule
        \Frozen$_\text{CC-clean}$ & 0.4M & 37.79 & \textbf{19.29} & \textbf{6.25} & \textbf{11.11} & 22.70 & 14.00 & 5.00 & 2.70 & \textbf{18.61} & 11.10 \\
        \Frozen$_\text{CC-cleanish}$ & 1.0M & \textbf{37.82} & 18.49 & 6.12 & 10.16 & 37.60 & 20.60 & 6.60 & 3.20 & 18.15 & 17.00 \\
        \Frozen$_\text{CC}$ & 2.7M & 37.81 & 18.33 & 5.56 & 9.97 & \textbf{57.60} & \textbf{22.20} & \textbf{8.00} & \textbf{4.20} & 17.92 & \textbf{23.00} \\
        \hdashline
        \Model$_\text{CC-clean}$ & 0.4M & 44.35 & 24.03 & \textbf{9.65} & 17.33 & 72.70 & \textbf{54.60} & \textbf{23.80} & \textbf{21.10} & 23.84 & 43.05 \\
        \Model$_\text{CC-cleanish}$ & 1.0M & \textbf{46.63} & \textbf{25.99} & 8.48 & \textbf{19.65} & 88.30 & 54.10 & 22.30 & 19.80 & \textbf{25.19} & \textbf{46.13} \\
        \Model$_\text{CC}$ & 2.7M & 43.26 & 20.96 & 5.20 & 19.31 & \textbf{101.10} & 44.10 & 16.70 & 15.90 & 22.18 & 44.45 \\
        \bottomrule
    \end{tabular}%
    }
    \vspace{-7pt}
    \caption{Impact of data quality and size. These experiments are run with one seed only.}
    \vspace{-15pt}
    \label{tab:noisy_data}
\end{table*}

\vspace{-2pt}
\subsection{Evaluation of domain-agnostic learning}
\label{sec:main_results}

We report few-shot VQA results in Table~\ref{tab:vqa} and image captioning results in Table~\ref{tab:caption}. Subscripts in the first column denote the training dataset. \textit{Overall} accuracies denote average of per-benchmark accuracies.
First, we see that \Model$_\text{CC-clean}$ substantially outperforms \Modelblind$_\text{CC-clean}$ both on VQA and image captioning, proving that the visual inputs are not ignored by the LM in \Model. Second, we find that \Model$_\text{CC-clean}$ outperforms \Frozen$_\text{CC-clean}$ by a considerable margin on all VL benchmarks (with overall accuracy improvements of +6.61\% 0-shot and +5.66\% 4-shot on VQA tasks, +4.35 BLEU@4 and +33.55 CIDEr on image captioning tasks). Importantly, this is achieved while training an order of magnitude fewer parameters (3.4M vs 40.3M). Next, \Model$_\text{CC-clean}$ is competitive compared to existing methods (MAGMA, VLKD, ClipCap) and concurrent work LiMBeR, despite training one-two orders of magnitude fewer parameters on significantly less multimodal data. Lastly, \Model$_\text{CC-clean}$'s performance is still far from the performance of Flamingo, which trains orders of magnitude more parameters on orders of magnitude more data. However, we believe \Model to be an effective method for scenarios with constrained computational resources. 
For \Model's qualitative results, see App.~\ref{sec:qualitative_results}.

\noindent\textbf{Low-data learning.}
When trained on only 1\% domain-agnostic data, \Model$_\text{CC-clean}$ outperforms \Frozen$_\text{CC-clean}$ for all image captioning evaluations (by +1.35 BLEU@4 and +8.23 CIDEr, overall) and all 0-shot VQA evaluations (by +2.60\% overall accuracy), while achieving competitive performance on 4- and 8-shot VQA evaluations. In summary, these results show the effectiveness of our method in low-data settings, highlighting its usefulness for applications where data is scarce.

\begin{table*}[t]
    \resizebox{\textwidth}{!}{%
    \begin{tabular}{lccc||cccccccccccc|ccc}
        \toprule
         & \textbf{Vision} & \textbf{Language} & \textbf{Mapping} & \multicolumn{2}{c}{\textbf{VQAv2}} & \multicolumn{2}{c}{\textbf{OK-VQA}} & \multicolumn{2}{c}{\textbf{TextVQA}} & \multicolumn{2}{c}{\textbf{VizWiz-VQA}} & \textbf{CC} & \textbf{COCO} & \textbf{TextCaps} & \textbf{VizWiz-Caps} & \multicolumn{3}{c}{\textbf{Overall}} \\
         & \textbf{encoder} & \textbf{model} & \textbf{network} & \textbf{0-shot} & \textbf{4-shot} & \textbf{0-shot} & \textbf{4-shot} & \textbf{0-shot} & \textbf{4-shot} & \textbf{0-shot} & \textbf{4-shot} & \textbf{CIDEr} & \textbf{CIDEr} & \textbf{CIDEr} & \textbf{CIDEr} & \textbf{0-shot} & \textbf{4-shot} & \textbf{CIDEr} \\
        \midrule
        \Frozen & NF-ResNet-50 & GPT-J & Transformer & 27.98 & 36.66 & 5.88 & 18.44 & 4.56 & 7.87 & 3.67 & 10.32 & 21.79 & 14.87 & 5.42 & 3.03 & 10.52 & 18.32 & 11.28 \\
        \Frozen & ViT-L/16 & GPT-J & Transformer & 27.82 & 36.60 & 5.64 & 16.26 & 3.67 & 5.27 & 4.70 & 11.77 & 13.19 & 8.77 & 2.88 & 2.35 & 10.46 & 17.48 & 6.80 \\
        \Frozen & ViT-L/14 & GPT-J & Transformer & 24.45 & 36.03 & 4.14 & 16.27 & 3.91 & 5.33 & 3.27 & 10.09 & 13.89 & 8.27 & 2.66 & 2.50 & 8.94 & 16.93 & 6.83 \\
        \hdashline
        \Model & CLIP-ViT-L/14 & GPT-J & Transformer & \textbf{33.54} & \textbf{45.13} & 13.84 & 24.25 & 8.26 & \textbf{8.88} & 11.72 & \textbf{18.46} & \textbf{79.75} & 54.30 & 22.95 & \textbf{20.95} & \textbf{16.84} & \textbf{24.18} & \textbf{44.49} \\
        \Model & IN-NF-ResNet-50 & GPT-J & Transformer & 28.12 & 40.86 & 10.86 & 21.66 & 6.15 & 7.01 & 6.40 & 14.22 & 39.64 & 32.99 & 12.41 & 9.55 & 12.88 & 20.94 & 23.65 \\
        \Model & IN-ViT-L/16 & GPT-J & Transformer & 31.70 & 43.75 & 11.13 & \textbf{25.50} & 6.16 & 7.40 & 8.93 & 16.45 & 56.33 & 45.80 & 17.12 & 16.28 & 14.48 & 23.28 & 33.88 \\
        \hline
        \Frozen & NF-ResNet-50 & OPT-6.7B & Transformer & 30.16 & 32.72 & 8.10 & 13.79 & 5.44 & 6.81 & 7.15 & 6.77 & 25.40 & 17.80 & 6.90 & 4.00 & 12.71 & 15.02 & 13.53 \\
        \Model & CLIP-ViT-L/14 & OPT-6.7B & Transformer & 23.26 & 33.95 & \textbf{15.27} & 16.25 & \textbf{8.90} & 6.41 & \textbf{15.40} & 9.47 & 66.60 & \textbf{54.40} & \textbf{23.30} & 19.60 & 15.71 & 16.52 & 40.98 \\
        \hline
        \Model & CLIP-ViT-L/14 & GPT-J & Linear & 30.55 & 37.09 & 12.20 & 16.69 & 7.02 & 5.80 & 8.81 & 12.49 & 60.00 & 43.80 & 18.30 & 13.70 & 14.65 & 18.02 & 33.95 \\
        \Model & CLIP-ViT-L/14 & GPT-J & MLP & 28.99 & 43.69 & 11.07 & 25.33 & 6.60 & 8.39 & 9.73 & 17.14 & 70.40 & 49.10 & 20.90 & 20.30 & 14.10 & 23.64 & 40.18 \\
        \bottomrule
    \end{tabular}%
    }
    \vspace{-7pt}
    \caption{Ablation studies. We assess the impact of the choice of vision encoder (top), LM (middle) and mapping network architecture (bottom). All models are trained on 100\% of CC-clean with a single seed. IN stands for ImageNet pre-training.}
    \vspace{-15pt}
    \label{tab:ablations2}
\end{table*}

\vspace{-2pt}
\subsection{Evaluation of in-domain learning}
\label{sec:indomain_results}

In Tables \ref{tab:vqa} and \ref{tab:caption}, we observe that both \Model and \Frozen benefit from directly training on in-domain data, compared to few-shot transfer from large-scale domain-agnostic pretraining. For instance, \Model$_\text{COCO}$ and \Frozen$_\text{COCO}$ respectively outperform \Model$_\text{CC-clean}$ and \Frozen$_\text{CC-clean}$ on VQAv2, OK-VQA and COCO Captions when trained on 100\% of data. Interestingly, this performance gap is larger for \Model compared to \Frozen by +2\% 0-shot accuracy and +3.77\% 4-shot accuracy averaged across VQAv2 and OK-VQA, and +9.35 BLEU@4 and +23.45 CIDEr on COCO Captions. A similar trend can be observed for TextCaps and TextVQA. Surprisingly, for 0-shot VQA and image captioning, training on just 1\% of in-domain data outperforms 100\% CC training for all benchmarks (except VizWiz-VQA) and both models. These results demonstrate the benefits of in-domain learning. When comparing \Model vs. \Frozen, we observe that \Model outperforms \Frozen for all tasks and benchmarks (except VizWiz-VQA) under both 100\% and 1\% in-domain settings. In fact, \Model trained on just 1\% in-domain data outperforms \Frozen trained on 100\% in-domain data by +3.41\% 0-shot accuracy and +1.14\% 4-shot accuracy averaged across VQAv2, OK-VQA and TextVQA. Thus, \Model is more effective than \Frozen at in-domain learning.

Contrary to the above trends, we observe that \Model$_\text{VizWiz}$ under 1\% in-domain training performs worse than \Model~$_\text{COCO}$ or \Model~$_\text{TextCaps}$ when evaluated on VizWiz-VQA. We hypothesize the visual embeddings extracted from CLIP's vision encoder for VizWiz images are not as good as those for COCO or TextCaps' images because the distribution of images in VizWiz (captured by visually-impaired people) is rather different from the distribution of images CLIP is trained on (scraped from the web), whereas for COCO and TextCaps this isn’t the case. When training \Model's mapping network on only 1\% of VizWiz data, we believe the data is not large enough to compensate for the OOD pretrained vision encoder, so \Model trained on COCO/TextCaps performs better on VizWiz-VQA. For in-domain training with 100\% of data and 4/8-shot VQA, the mapping network has enough data to learn from and compensate for the OOD phenomenon. On the other hand, \Frozen does not suffer from this issue because its vision encoder is trained from scratch, allowing it to adapt to the image distribution.

Lastly, we observe that \Model$_\text{COCO}$ outperforms ClipCap\xspace~$_\text{COCO}$ by +2.92 BLEU@4 and +12.12 CIDEr on COCO Captions. \Model$_\text{COCO}$ also outperforms \PICa (which represents images with ground-truth COCO captions) on VQAv2 and 0-shot OK-VQA, and achieves competitive results on few-shot OK-VQA; this demonstrates representing images with continuous embeddings is beneficial over caption-based image representations. Overall, we see that in-domain learning is beneficial and \Model is more effective at it than similar methods.

\vspace{-2pt}
\subsection{Impact of data quality and size}
\label{sec:noisy_data}

To measure the impact of noise in the training data, we additionally train \Model and \Frozen on the \textit{full} CC dataset, consisting of 2.8M\footnote{This is not the full 3.3M CC due to broken URLs.} examples, as well as on a \textit{clean-ish} version consisting of the 1.0M most similar image-text pairs. In Table~\ref{tab:noisy_data}, we observe \Frozen achieves similar performance on few-shot VQA tasks when trained on noisy vs. clean data; however, \Frozen's performance on image captioning decreases when trained on cleaner but smaller data. In contrast, \Model generally benefits from cleaner training data, with the exception of evaluation on CC. We hypothesize both models perform better on CC when trained on larger (yet noisier) data because the CC validation set is IID with the full (noisy) CC training set. In the case of \Frozen, as we move away from the IID setting, the benefits from more data start diminishing (CC captioning > other captioning tasks > VQA tasks). For \Model, the benefit from reduced noise in training data exceeds the degradation caused by a smaller data size, thanks to the reduced number of trainable parameters. These trends align with previous observations that larger models are more robust to noisy training data since they have enough capacity to model both noise and the desired function~\cite{rolnick2017deep}, while smaller models are more sample-efficient \cite{vapnik2015uniform}, i.e. they need less (clean) data to train effectively. Note that although \Model's overall performance is higher when training on 1.0M than on 0.4M examples, we decided to train with 0.4M examples because training on ${\sim}2.5\times$ more data (1.0M instead of 0.4M) required ${\sim}5\times$ more iterations (always early-stopping based on validation loss). So we did not think the slight performance increase due to more data was worth the ${\sim}5\times$ longer training time, especially because we were operating under a limited compute budget.

\vspace{-2pt}
\subsection{Ablation studies}
\label{sec:ablations}

In this section, we evaluate how the choice of vision encoder, LM and mapping network architecture impact \Model's performance, and compare it with corresponding versions of \Frozen (where applicable). Results are presented in Table~\ref{tab:ablations2}. Please refer to App.~\ref{sec:more_ablations} for more ablations.

First, to assess the impact of the choice of vision encoder, we train additional versions of \Model replacing the CLIP pre-trained vision encoder (ViT-L/14 -- 303M parameters) with encoders pre-trained on ImageNet: NF-ResNet-50 (23.5M) and ViT-L/16 (303M), and compare their performance with corresponding versions of \Frozen. We observe that: 1) \Model outperforms \Frozen for each configuration of vision encoder, suggesting that \Model is \textbf{robust to the choice of vision encoder's pre-training data and architecture}; and 2) \Frozen’s performance drops with bigger vision encoders (likely due to more trainable parameters), whereas \Model improves due to the use of stronger pre-trained encoders. Thus, training the vision encoder from scratch (\Frozen) has limited application, while \Model's \textbf{performance scales alongside the pre-trained vision encoder}.

Next, to evaluate the impact of the choice of LM, we train both \Model and \Frozen replacing GPT-J by OPT-6.7B~\cite{zhang2022opt}. We see that in all settings except 0-shot VQAv2, \Model outperforms \Frozen. See App.~\ref{sec:0shot_vqav2_opt} for discussion on 0-shot VQAv2 results. This suggests that \Model is \textbf{robust to the choice of LM}. The above results also highlight how \Model's \textbf{modularity} allows to easily replace the pre-trained vision encoder or the LM.\looseness=-1

Lastly, to assess the impact of the choice of mapping network architecture, we replace the proposed transformer-based mapping network with two simpler architectures -- a linear layer and a 2-layer MLP (see App.~\ref{sec:simple_mappings_details} for details). We observe both these versions generally underperform the original setting (transformer-based), highlighting the \textbf{effectiveness of the proposed design}. We also note that in these simpler versions, the parameter count is directly proportional to the vision encoder's representation size and LM's embedding size, whereas in \Model we decouple this using a dimensionality bottleneck (Sec.~\ref{sec:architecture}), making our mapping network \textbf{more parameter-efficient by design}.

\vspace{-2pt}
\subsection{Qualitative results}
\label{sec:qualitative_results}
\begin{figure}[ht]
    \centering
    \includegraphics[width=\linewidth]{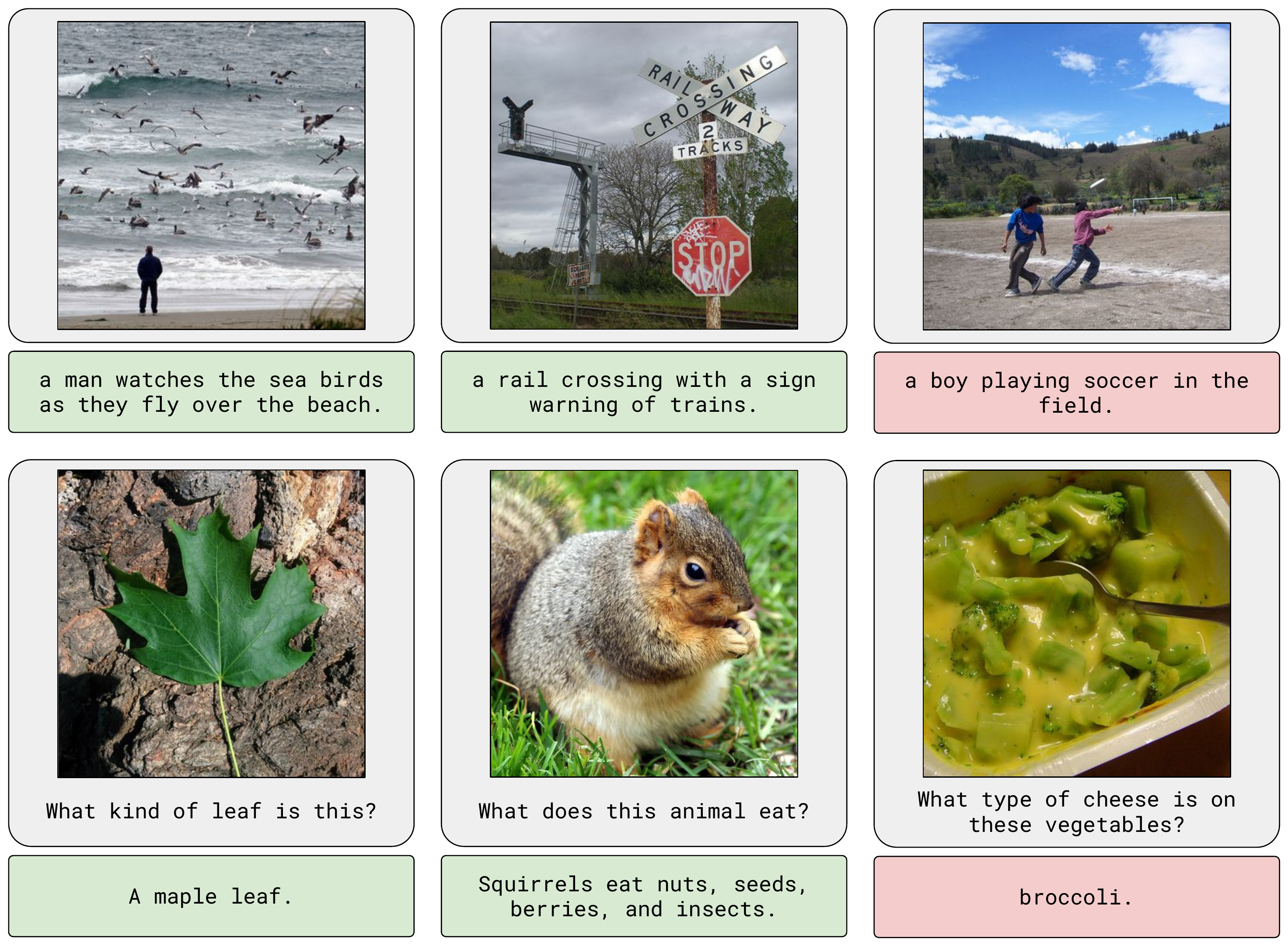}
    \caption{Qualitative samples from the web using \Model$_\text{CC-clean}$. (Multimodal) input is in gray, and \Model's output is in green (success) or red (failure).}
    \vspace{-15pt}
    \label{fig:qualitative_samples}
\end{figure}

Figure~\ref{fig:qualitative_samples} shows some selected samples from the web illustrating our interface at inference time using \Model$_\text{CC-clean}$. The first two columns show successful results while the last column shows failure cases. For image captioning (top row), success cases show \Model can generate meaningful and detailed textual descriptions of the scene. For zero-shot VQA (bottom row), success cases indicate that \Model is able to parse the question and connect visual information to encyclopedic knowledge contained in the pre-trained LM. However, \Model's visio-linguistic understanding is evidently still far from being perfect. More qualitative results (both success and failure cases) are provided in App.~\ref{sec:more_qualitative}.

\section{Conclusion}
We introduce \Model, a simple and parameter-efficient method to repurpose pre-trained and frozen unimodal models for multimodal tasks. Our experiments demonstrate that \Model achieves superior or competitive performance compared to similar methods on several VL benchmarks while training orders of magnitude fewer parameters. Importantly, we also show that \Model is effective in the low-data and in-domain settings thanks to its reduced number of trainable parameters. We leave as future work exploring training on a weighted mixture of image-text datasets, evaluating on more downstream tasks such as NLVR2~\cite{suhr2019corpus} and Visual Dialog~\cite{das2017visual}, and investigating the use of masked LMs~\cite{schick2021s,chung2022scaling} with \Model.

\section*{Limitations}
\Model achieves reasonable performance on VL tasks, but it is still far from the performance of recent methods leveraging large-scale data and compute. On the other hand, \Model is a preferable alternative in scenarios with constrained computational resources.

We observed our mapping network is sensitive to initialization, so different random seeds can yield non-negligible variance in downstream performance. We think this might be related to the reduced number of trainable parameters. We tried to reduce the effect of this variance by reporting average performance across different seeds. We also observed \Model struggles to leverage more shots for in-context learning. We hypothesize this could be caused by our model being trained on single image-caption pairs -- as opposed to the sequences of multiple images and texts seen during few-shot transfer, so a better pretext task might help (see App.~\ref{sec:more_shots} for further discussion).

\Model builds on top of pre-trained vision-only and language-only models, inheriting their capabilities but also their limitations. An important risk is that our model might inherit the existing social, gender or racial biases of pre-trained models. However, our limited qualitative analysis (see App.~\ref{sec:qualitative_analysis}) shows that providing visual information significantly changes the prior answer distribution of the LM. Therefore, how much of the underlying bias is retained remains an empirical question.

\section*{Ethics Statement}
\paragraph{Model recycling.}
\Model reuses vision-only and language-only foundation models. Hence, the expensive computational resources used to train these models can be amortized to help reduce energy and carbon costs.

\paragraph{Public datasets.}
\Model is trained uniquely on publicly available datasets, which facilitates reproducibility and provides transparency on the origin and the characteristics of the data the model has seen.

\paragraph{Undesired biases.}
\Model could be exposed to undesired biases from different sources. The pre-trained vision encoder might have been trained with data where certain races or genders are underrepresented, hence biasing our representation of images. The pre-trained LM might also be biased towards generating toxic or offensive language when fed with certain prompts. Finally, the image-text data used to align the representation spaces of such models was annotated by humans, so it might reflect a biased view of the world.

\paragraph{Broader impact.}
This work shows how one can easily adapt pre-trained vision encoders and LMs for multimodal tasks. Given the parameter-efficiency of our method, we believe it should be of great interest to the sections of the community that do not have access to large compute resources (e.g., small academic labs and independent researchers), and for low-data applications. While \Model can be applied in many useful applications (e.g., aiding visually-impaired people), it also makes it simpler to create malicious or offensive multimodal systems from existing unimodal models. Further research efforts are needed on how to safely deploy such systems so that their behavior always aligns with ethical values.

\section*{Acknowledgements}
We thank Pauline Luc, Siva Reddy, Lisa Anne Hendricks and Chris Dyer for their constructive feedback. We are grateful to the Mila IDT team for their technical support with the computational infrastructure. During this project, Aishwarya Agrawal was supported by the Canada CIFAR AI Chair award. We would also like to thank Samsung Electronics Co., Ldt. for funding this research.

\bibliography{references}

\begin{thebibliography}{50}
\expandafter\ifx\csname natexlab\endcsname\relax\def\natexlab#1{#1}\fi

\bibitem[{Agrawal et~al.(2022)Agrawal, Kaji{\'c}, Bugliarello, Davoodi,
  Gergely, Blunsom, and Nematzadeh}]{agrawal2022rethinking}
Aishwarya Agrawal, Ivana Kaji{\'c}, Emanuele Bugliarello, Elnaz Davoodi, Anita
  Gergely, Phil Blunsom, and Aida Nematzadeh. 2022.
\newblock Rethinking evaluation practices in visual question answering: A case
  study on out-of-distribution generalization.
\newblock \emph{arXiv preprint arXiv:2205.12191}.

\bibitem[{Alayrac et~al.(2022)Alayrac, Donahue, Luc, Miech, Barr, Hasson, Lenc,
  Mensch, Millican, Reynolds et~al.}]{alayrac2022flamingo}
Jean-Baptiste Alayrac, Jeff Donahue, Pauline Luc, Antoine Miech, Iain Barr,
  Yana Hasson, Karel Lenc, Arthur Mensch, Katie Millican, Malcolm Reynolds,
  et~al. 2022.
\newblock Flamingo: a visual language model for few-shot learning.
\newblock \emph{arXiv preprint arXiv:2204.14198}.

\bibitem[{Antol et~al.(2015)Antol, Agrawal, Lu, Mitchell, Batra, Zitnick, and
  Parikh}]{antol2015vqa}
Stanislaw Antol, Aishwarya Agrawal, Jiasen Lu, Margaret Mitchell, Dhruv Batra,
  C~Lawrence Zitnick, and Devi Parikh. 2015.
\newblock Vqa: Visual question answering.
\newblock In \emph{Proceedings of the IEEE international conference on computer
  vision}, pages 2425--2433.

\bibitem[{Bommasani et~al.(2021)Bommasani, Hudson, Adeli, Altman, Arora, von
  Arx, Bernstein, Bohg, Bosselut, Brunskill
  et~al.}]{bommasani2021opportunities}
Rishi Bommasani, Drew~A Hudson, Ehsan Adeli, Russ Altman, Simran Arora, Sydney
  von Arx, Michael~S Bernstein, Jeannette Bohg, Antoine Bosselut, Emma
  Brunskill, et~al. 2021.
\newblock On the opportunities and risks of foundation models.
\newblock \emph{arXiv preprint arXiv:2108.07258}.

\bibitem[{Brown et~al.(2020)Brown, Mann, Ryder, Subbiah, Kaplan, Dhariwal,
  Neelakantan, Shyam, Sastry, Askell et~al.}]{brown2020language}
Tom Brown, Benjamin Mann, Nick Ryder, Melanie Subbiah, Jared~D Kaplan, Prafulla
  Dhariwal, Arvind Neelakantan, Pranav Shyam, Girish Sastry, Amanda Askell,
  et~al. 2020.
\newblock Language models are few-shot learners.
\newblock \emph{Advances in neural information processing systems},
  33:1877--1901.

\bibitem[{Carion et~al.(2020)Carion, Massa, Synnaeve, Usunier, Kirillov, and
  Zagoruyko}]{carion2020end}
Nicolas Carion, Francisco Massa, Gabriel Synnaeve, Nicolas Usunier, Alexander
  Kirillov, and Sergey Zagoruyko. 2020.
\newblock End-to-end object detection with transformers.
\newblock In \emph{European conference on computer vision}, pages 213--229.
  Springer.

\bibitem[{Chen et~al.(2021)Chen, Guo, Yi, Li, and
  Elhoseiny}]{chen2021visualgpt}
Jun Chen, Han Guo, Kai Yi, Boyang Li, and Mohamed Elhoseiny. 2021.
\newblock Visualgpt: Data-efficient adaptation of pretrained language models
  for image captioning.
\newblock \emph{arXiv preprint arXiv:2102.10407}.

\bibitem[{Chen et~al.(2015)Chen, Fang, Lin, Vedantam, Gupta, Doll{\'a}r, and
  Zitnick}]{chen2015microsoft}
Xinlei Chen, Hao Fang, Tsung-Yi Lin, Ramakrishna Vedantam, Saurabh Gupta, Piotr
  Doll{\'a}r, and C~Lawrence Zitnick. 2015.
\newblock Microsoft coco captions: Data collection and evaluation server.
\newblock \emph{arXiv preprint arXiv:1504.00325}.

\bibitem[{Chen et~al.(2019)Chen, Li, Yu, El~Kholy, Ahmed, Gan, Cheng, and
  Liu}]{chen2019uniter}
Yen-Chun Chen, Linjie Li, Licheng Yu, Ahmed El~Kholy, Faisal Ahmed, Zhe Gan,
  Yu~Cheng, and Jingjing Liu. 2019.
\newblock Uniter: Learning universal image-text representations.

\bibitem[{Cho et~al.(2021)Cho, Lei, Tan, and Bansal}]{cho2021unifying}
Jaemin Cho, Jie Lei, Hao Tan, and Mohit Bansal. 2021.
\newblock Unifying vision-and-language tasks via text generation.
\newblock In \emph{International Conference on Machine Learning}, pages
  1931--1942. PMLR.

\bibitem[{Chung et~al.(2022)Chung, Hou, Longpre, Zoph, Tay, Fedus, Li, Wang,
  Dehghani, Brahma et~al.}]{chung2022scaling}
Hyung~Won Chung, Le~Hou, Shayne Longpre, Barret Zoph, Yi~Tay, William Fedus,
  Eric Li, Xuezhi Wang, Mostafa Dehghani, Siddhartha Brahma, et~al. 2022.
\newblock Scaling instruction-finetuned language models.
\newblock \emph{arXiv preprint arXiv:2210.11416}.

\bibitem[{Dai et~al.(2022)Dai, Hou, Shang, Jiang, Liu, and
  Fung}]{dai2022enabling}
Wenliang Dai, Lu~Hou, Lifeng Shang, Xin Jiang, Qun Liu, and Pascale Fung. 2022.
\newblock \href {https://doi.org/10.18653/v1/2022.findings-acl.187} {Enabling
  multimodal generation on {CLIP} via vision-language knowledge distillation}.
\newblock In \emph{Findings of the Association for Computational Linguistics:
  ACL 2022}, pages 2383--2395, Dublin, Ireland. Association for Computational
  Linguistics.

\bibitem[{Das et~al.(2017)Das, Kottur, Gupta, Singh, Yadav, Moura, Parikh, and
  Batra}]{das2017visual}
Abhishek Das, Satwik Kottur, Khushi Gupta, Avi Singh, Deshraj Yadav,
  Jos{\'e}~MF Moura, Devi Parikh, and Dhruv Batra. 2017.
\newblock Visual dialog.
\newblock In \emph{Proceedings of the IEEE conference on computer vision and
  pattern recognition}, pages 326--335.

\bibitem[{Dosovitskiy et~al.(2020)Dosovitskiy, Beyer, Kolesnikov, Weissenborn,
  Zhai, Unterthiner, Dehghani, Minderer, Heigold, Gelly
  et~al.}]{dosovitskiy2020image}
Alexey Dosovitskiy, Lucas Beyer, Alexander Kolesnikov, Dirk Weissenborn,
  Xiaohua Zhai, Thomas Unterthiner, Mostafa Dehghani, Matthias Minderer, Georg
  Heigold, Sylvain Gelly, et~al. 2020.
\newblock An image is worth 16x16 words: Transformers for image recognition at
  scale.
\newblock In \emph{ICLR}.

\bibitem[{Eichenberg et~al.(2021)Eichenberg, Black, Weinbach, Parcalabescu, and
  Frank}]{eichenberg2021magma}
Constantin Eichenberg, Sidney Black, Samuel Weinbach, Letitia Parcalabescu, and
  Anette Frank. 2021.
\newblock Magma--multimodal augmentation of generative models through
  adapter-based finetuning.
\newblock \emph{arXiv preprint arXiv:2112.05253}.

\bibitem[{Gao et~al.(2020)Gao, Biderman, Black, Golding, Hoppe, Foster, Phang,
  He, Thite, Nabeshima et~al.}]{gao2020pile}
Leo Gao, Stella Biderman, Sid Black, Laurence Golding, Travis Hoppe, Charles
  Foster, Jason Phang, Horace He, Anish Thite, Noa Nabeshima, et~al. 2020.
\newblock The pile: An 800gb dataset of diverse text for language modeling.
\newblock \emph{arXiv preprint arXiv:2101.00027}.

\bibitem[{Goyal et~al.(2017)Goyal, Khot, Summers-Stay, Batra, and
  Parikh}]{goyal2017making}
Yash Goyal, Tejas Khot, Douglas Summers-Stay, Dhruv Batra, and Devi Parikh.
  2017.
\newblock Making the v in vqa matter: Elevating the role of image understanding
  in visual question answering.
\newblock In \emph{Proceedings of the IEEE conference on computer vision and
  pattern recognition}, pages 6904--6913.

\bibitem[{Gurari et~al.(2018)Gurari, Li, Stangl, Guo, Lin, Grauman, Luo, and
  Bigham}]{gurari2018vizwiz}
Danna Gurari, Qing Li, Abigale~J Stangl, Anhong Guo, Chi Lin, Kristen Grauman,
  Jiebo Luo, and Jeffrey~P Bigham. 2018.
\newblock Vizwiz grand challenge: Answering visual questions from blind people.
\newblock In \emph{Proceedings of the IEEE Conference on Computer Vision and
  Pattern Recognition}, pages 3608--3617.

\bibitem[{Hao et~al.(2022)Hao, Song, Dong, Huang, Chi, Wang, Ma, and
  Wei}]{hao2022language}
Yaru Hao, Haoyu Song, Li~Dong, Shaohan Huang, Zewen Chi, Wenhui Wang, Shuming
  Ma, and Furu Wei. 2022.
\newblock Language models are general-purpose interfaces.
\newblock \emph{arXiv preprint arXiv:2206.06336}.

\bibitem[{Jia et~al.(2021)Jia, Yang, Xia, Chen, Parekh, Pham, Le, Sung, Li, and
  Duerig}]{jia2021scaling}
Chao Jia, Yinfei Yang, Ye~Xia, Yi-Ting Chen, Zarana Parekh, Hieu Pham, Quoc Le,
  Yun-Hsuan Sung, Zhen Li, and Tom Duerig. 2021.
\newblock Scaling up visual and vision-language representation learning with
  noisy text supervision.
\newblock In \emph{International Conference on Machine Learning}, pages
  4904--4916. PMLR.

\bibitem[{Jin et~al.(2022)Jin, Cheng, Shen, Chen, and Ren}]{jin2021good}
Woojeong Jin, Yu~Cheng, Yelong Shen, Weizhu Chen, and Xiang Ren. 2022.
\newblock \href {https://doi.org/10.18653/v1/2022.acl-long.197} {A good prompt
  is worth millions of parameters: Low-resource prompt-based learning for
  vision-language models}.
\newblock In \emph{Proceedings of the 60th Annual Meeting of the Association
  for Computational Linguistics (Volume 1: Long Papers)}, pages 2763--2775,
  Dublin, Ireland. Association for Computational Linguistics.

\bibitem[{Karpathy and Fei-Fei(2015)}]{karpathy2015deep}
Andrej Karpathy and Li~Fei-Fei. 2015.
\newblock Deep visual-semantic alignments for generating image descriptions.
\newblock In \emph{Proceedings of the IEEE conference on computer vision and
  pattern recognition}, pages 3128--3137.

\bibitem[{Lamb et~al.(2016)Lamb, ALIAS PARTH~GOYAL, Zhang, Zhang, Courville,
  and Bengio}]{lamb2016professor}
Alex~M Lamb, Anirudh~Goyal ALIAS PARTH~GOYAL, Ying Zhang, Saizheng Zhang,
  Aaron~C Courville, and Yoshua Bengio. 2016.
\newblock Professor forcing: A new algorithm for training recurrent networks.
\newblock \emph{Advances in neural information processing systems}, 29.

\bibitem[{Li et~al.(2021)Li, Selvaraju, Gotmare, Joty, Xiong, and
  Hoi}]{li2021align}
Junnan Li, Ramprasaath Selvaraju, Akhilesh Gotmare, Shafiq Joty, Caiming Xiong,
  and Steven Chu~Hong Hoi. 2021.
\newblock Align before fuse: Vision and language representation learning with
  momentum distillation.
\newblock \emph{Advances in Neural Information Processing Systems}, 34.

\bibitem[{Li et~al.(2020)Li, Yin, Li, Zhang, Hu, Zhang, Wang, Hu, Dong, Wei
  et~al.}]{li2020oscar}
Xiujun Li, Xi~Yin, Chunyuan Li, Pengchuan Zhang, Xiaowei Hu, Lei Zhang, Lijuan
  Wang, Houdong Hu, Li~Dong, Furu Wei, et~al. 2020.
\newblock Oscar: Object-semantics aligned pre-training for vision-language
  tasks.
\newblock In \emph{European Conference on Computer Vision}, pages 121--137.
  Springer.

\bibitem[{Lu et~al.(2019)Lu, Batra, Parikh, and Lee}]{lu2019vilbert}
Jiasen Lu, Dhruv Batra, Devi Parikh, and Stefan Lee. 2019.
\newblock Vilbert: Pretraining task-agnostic visiolinguistic representations
  for vision-and-language tasks.
\newblock \emph{Advances in neural information processing systems}, 32.

\bibitem[{Luo et~al.(2022)Luo, Xi, Zhang, and
  Ma}]{DBLP:journals/corr/abs-2201-12723}
Ziyang Luo, Yadong Xi, Rongsheng Zhang, and Jing Ma. 2022.
\newblock {VC-GPT:} visual conditioned {GPT} for end-to-end generative
  vision-and-language pre-training.
\newblock \emph{CoRR}, abs/2201.12723.

\bibitem[{Marino et~al.(2019)Marino, Rastegari, Farhadi, and
  Mottaghi}]{marino2019ok}
Kenneth Marino, Mohammad Rastegari, Ali Farhadi, and Roozbeh Mottaghi. 2019.
\newblock Ok-vqa: A visual question answering benchmark requiring external
  knowledge.
\newblock In \emph{Proceedings of the IEEE/CVF Conference on Computer Vision
  and Pattern Recognition}, pages 3195--3204.

\bibitem[{Merullo et~al.(2022)Merullo, Castricato, Eickhoff, and
  Pavlick}]{merullo2022linearly}
Jack Merullo, Louis Castricato, Carsten Eickhoff, and Ellie Pavlick. 2022.
\newblock Linearly mapping from image to text space.
\newblock \emph{arXiv preprint arXiv:2209.15162}.

\bibitem[{Mokady et~al.(2021)Mokady, Hertz, and Bermano}]{mokady2021clipcap}
Ron Mokady, Amir Hertz, and Amit~H Bermano. 2021.
\newblock Clipcap: Clip prefix for image captioning.
\newblock \emph{arXiv preprint arXiv:2111.09734}.

\bibitem[{Radford et~al.(2021)Radford, Kim, Hallacy, Ramesh, Goh, Agarwal,
  Sastry, Askell, Mishkin, Clark et~al.}]{radford2021learning}
Alec Radford, Jong~Wook Kim, Chris Hallacy, Aditya Ramesh, Gabriel Goh,
  Sandhini Agarwal, Girish Sastry, Amanda Askell, Pamela Mishkin, Jack Clark,
  et~al. 2021.
\newblock Learning transferable visual models from natural language
  supervision.
\newblock In \emph{International Conference on Machine Learning}, pages
  8748--8763. PMLR.

\bibitem[{Radford et~al.(2018)Radford, Narasimhan, Salimans, and
  Sutskever}]{radford2018improving}
Alec Radford, Karthik Narasimhan, Tim Salimans, and Ilya Sutskever. 2018.
\newblock Improving language understanding by generative pre-training.

\bibitem[{Radford et~al.(2019)Radford, Wu, Child, Luan, Amodei, Sutskever
  et~al.}]{radford2019language}
Alec Radford, Jeffrey Wu, Rewon Child, David Luan, Dario Amodei, Ilya
  Sutskever, et~al. 2019.
\newblock Language models are unsupervised multitask learners.
\newblock \emph{OpenAI blog}, 1(8):9.

\bibitem[{Rajbhandari et~al.(2020)Rajbhandari, Rasley, Ruwase, and
  He}]{rajbhandari2020zero}
Samyam Rajbhandari, Jeff Rasley, Olatunji Ruwase, and Yuxiong He. 2020.
\newblock Zero: Memory optimizations toward training trillion parameter models.
\newblock In \emph{SC20: International Conference for High Performance
  Computing, Networking, Storage and Analysis}, pages 1--16. IEEE.

\bibitem[{Reynolds and McDonell(2021)}]{reynolds2021prompt}
Laria Reynolds and Kyle McDonell. 2021.
\newblock Prompt programming for large language models: Beyond the few-shot
  paradigm.
\newblock In \emph{Extended Abstracts of the 2021 CHI Conference on Human
  Factors in Computing Systems}, pages 1--7.

\bibitem[{Rolnick et~al.(2017)Rolnick, Veit, Belongie, and
  Shavit}]{rolnick2017deep}
David Rolnick, Andreas Veit, Serge Belongie, and Nir Shavit. 2017.
\newblock Deep learning is robust to massive label noise.
\newblock \emph{arXiv preprint arXiv:1705.10694}.

\bibitem[{Schick and Sch{\"u}tze(2021)}]{schick2021s}
Timo Schick and Hinrich Sch{\"u}tze. 2021.
\newblock It’s not just size that matters: Small language models are also
  few-shot learners.
\newblock In \emph{Proceedings of the 2021 Conference of the North American
  Chapter of the Association for Computational Linguistics: Human Language
  Technologies}, pages 2339--2352.

\bibitem[{Sharma et~al.(2018)Sharma, Ding, Goodman, and
  Soricut}]{sharma2018conceptual}
Piyush Sharma, Nan Ding, Sebastian Goodman, and Radu Soricut. 2018.
\newblock Conceptual captions: A cleaned, hypernymed, image alt-text dataset
  for automatic image captioning.
\newblock In \emph{Proceedings of the 56th Annual Meeting of the Association
  for Computational Linguistics (Volume 1: Long Papers)}, pages 2556--2565.

\bibitem[{Shaw et~al.(2018)Shaw, Uszkoreit, and Vaswani}]{shaw2018self}
Peter Shaw, Jakob Uszkoreit, and Ashish Vaswani. 2018.
\newblock Self-attention with relative position representations.
\newblock In \emph{NAACL-HLT (2)}.

\bibitem[{Sidorov et~al.(2020)Sidorov, Hu, Rohrbach, and
  Singh}]{sidorov2020textcaps}
Oleksii Sidorov, Ronghang Hu, Marcus Rohrbach, and Amanpreet Singh. 2020.
\newblock Textcaps: a dataset for image captioning with reading comprehension.
\newblock In \emph{European Conference on Computer Vision}, pages 742--758.
  Springer.

\bibitem[{Singh et~al.(2019)Singh, Natarajan, Shah, Jiang, Chen, Batra, Parikh,
  and Rohrbach}]{singh2019towards}
Amanpreet Singh, Vivek Natarajan, Meet Shah, Yu~Jiang, Xinlei Chen, Dhruv
  Batra, Devi Parikh, and Marcus Rohrbach. 2019.
\newblock Towards vqa models that can read.
\newblock In \emph{Proceedings of the IEEE/CVF Conference on Computer Vision
  and Pattern Recognition}, pages 8317--8326.

\bibitem[{Suhr et~al.(2019)Suhr, Zhou, Zhang, Zhang, Bai, and
  Artzi}]{suhr2019corpus}
Alane Suhr, Stephanie Zhou, Ally Zhang, Iris Zhang, Huajun Bai, and Yoav Artzi.
  2019.
\newblock A corpus for reasoning about natural language grounded in
  photographs.
\newblock In \emph{Proceedings of the 57th Annual Meeting of the Association
  for Computational Linguistics}, pages 6418--6428.

\bibitem[{Tan and Bansal(2019)}]{tan2019lxmert}
Hao Tan and Mohit Bansal. 2019.
\newblock Lxmert: Learning cross-modality encoder representations from
  transformers.
\newblock In \emph{Proceedings of the 2019 Conference on Empirical Methods in
  Natural Language Processing and the 9th International Joint Conference on
  Natural Language Processing (EMNLP-IJCNLP)}, pages 5100--5111.

\bibitem[{Tsimpoukelli et~al.(2021)Tsimpoukelli, Menick, Cabi, Eslami, Vinyals,
  and Hill}]{tsimpoukelli2021multimodal}
Maria Tsimpoukelli, Jacob Menick, Serkan Cabi, SM~Eslami, Oriol Vinyals, and
  Felix Hill. 2021.
\newblock Multimodal few-shot learning with frozen language models.
\newblock \emph{Advances in Neural Information Processing Systems}, 34.

\bibitem[{Vapnik and Chervonenkis(2015)}]{vapnik2015uniform}
Vladimir~N Vapnik and A~Ya Chervonenkis. 2015.
\newblock On the uniform convergence of relative frequencies of events to their
  probabilities.
\newblock In \emph{Measures of complexity}, pages 11--30. Springer.

\bibitem[{Wang and Komatsuzaki(2021)}]{gpt-j}
Ben Wang and Aran Komatsuzaki. 2021.
\newblock {GPT-J-6B: A 6 Billion Parameter Autoregressive Language Model}.
\newblock \url{https://github.com/kingoflolz/mesh-transformer-jax}.

\bibitem[{Wang et~al.(2021)Wang, Yu, Yu, Dai, Tsvetkov, and
  Cao}]{wang2021simvlm}
Zirui Wang, Jiahui Yu, Adams~Wei Yu, Zihang Dai, Yulia Tsvetkov, and Yuan Cao.
  2021.
\newblock Simvlm: Simple visual language model pretraining with weak
  supervision.
\newblock \emph{arXiv preprint arXiv:2108.10904}.

\bibitem[{Yang et~al.(2021)Yang, Gan, Wang, Hu, Lu, Liu, and
  Wang}]{yang2021empirical}
Zhengyuan Yang, Zhe Gan, Jianfeng Wang, Xiaowei Hu, Yumao Lu, Zicheng Liu, and
  Lijuan Wang. 2021.
\newblock An empirical study of gpt-3 for few-shot knowledge-based vqa.
\newblock \emph{arXiv preprint arXiv:2109.05014}.

\bibitem[{Zhang et~al.(2021)Zhang, Li, Hu, Yang, Zhang, Wang, Choi, and
  Gao}]{zhang2021vinvl}
Pengchuan Zhang, Xiujun Li, Xiaowei Hu, Jianwei Yang, Lei Zhang, Lijuan Wang,
  Yejin Choi, and Jianfeng Gao. 2021.
\newblock Vinvl: Revisiting visual representations in vision-language models.
\newblock In \emph{Proceedings of the IEEE/CVF Conference on Computer Vision
  and Pattern Recognition}, pages 5579--5588.

\bibitem[{Zhang et~al.(2022)Zhang, Roller, Goyal, Artetxe, Chen, Chen, Dewan,
  Diab, Li, Lin et~al.}]{zhang2022opt}
Susan Zhang, Stephen Roller, Naman Goyal, Mikel Artetxe, Moya Chen, Shuohui
  Chen, Christopher Dewan, Mona Diab, Xian Li, Xi~Victoria Lin, et~al. 2022.
\newblock Opt: Open pre-trained transformer language models.
\newblock \emph{arXiv preprint arXiv:2205.01068}.

\end{thebibliography}
\bibliographystyle{acl_natbib}

\clearpage
\appendix
\section{Appendix}
\label{sec:appendix}
\begin{table*}[t]
    \resizebox{\textwidth}{!}{%
    \begin{tabular}{rlll||cccccccc|cc}
        \toprule
         & \textbf{Ablated} & \textbf{Original} & \textbf{Changed} & \textbf{VQAv2} & \textbf{OK-VQA} & \textbf{TextVQA} & \textbf{VizWiz-VQA} & \textbf{CC} & \textbf{COCO} & \textbf{TextCaps} & \textbf{VizWiz-Caps} & \multicolumn{2}{c}{\textbf{Overall $\Delta$}} \\
         & \textbf{setting} & \textbf{value} & \textbf{value} & \textbf{4-shot} & \textbf{4-shot} & \textbf{4-shot} & \textbf{4-shot} & \textbf{CIDEr} & \textbf{CIDEr} & \textbf{CIDEr} & \textbf{CIDEr} & \textbf{4-shot} & \textbf{CIDEr} \\
        \midrule
         & \multicolumn{3}{c||}{\Model} & 46.39 & 25.49 & 9.87 & 20.02 & 71.90 & 54.90 & 23.30 & 21.70 & 0 & 0 \\
        \hdashline
        \textbf{(i)} & Vision encoder & CLIP-ViT-L/14 & CLIP-ViT-B/32 & 43.48 & 24.43 & 7.85 & 15.97 & 59.20 & 47.00 & 19.00 & 15.70 & -2.51 & -7.73 \\
        \hdashline
        \textbf{(ii)} & Visual features & Grid & Global & 43.75 & 22.90 & 8.81 & 18.20 & 66.70 & 49.70 & 18.40 & 19.90 & -2.03 & -4.28 \\
        \hdashline
        \multirow{2}{*}{\textbf{(iii)}} & Mapping & \multirow{2}{*}{Medium} & Small & 44.83 & 26.37 & 9.68 & 17.78 & 68.30 & 55.50 & 21.30 & 20.80 & -0.78 & -1.48 \\
         & network size & & Large & 45.03 & 23.92 & 8.88 & 19.01 & 73.40 & 57.10 & 24.00 & 23.10 & -1.23 & +1.45 \\
        \hdashline
        \multirow{2}{*}{\textbf{(iv)}} & Output & \multirow{2}{*}{32} & 16 & 44.18 & 25.16 & 9.01 & 18.15 & 72.80 & 56.20 & 22.50 & 21.80 & -1.32 & +0.37 \\
         & seq. length & & 64 & 45.22 & 25.07 & 10.35 & 18.89 & 74.80 & 58.30 & 24.30 & 24.80 & -0.56 & +2.60 \\
        \hdashline
        \multirow{2}{*}{\textbf{(v)}} & Learned constant & \multirow{2}{*}{Yes} & \multirow{2}{*}{No} & \multirow{2}{*}{40.87} & \multirow{2}{*}{19.31} & \multirow{2}{*}{11.42} & \multirow{2}{*}{16.79} & \multirow{2}{*}{80.52} & \multirow{2}{*}{57.49} & \multirow{2}{*}{30.07} & \multirow{2}{*}{26.16} & \multirow{2}{*}{-3.35} & \multirow{2}{*}{+5.61} \\
         & embeddings & & & & & & & & & & & & \\
        \hdashline
        \multirow{2}{*}{\textbf{(vi)}} & Data quality & Clean & Noisy & \multirow{2}{*}{42.80} & \multirow{2}{*}{22.59} & \multirow{2}{*}{6.06} & \multirow{2}{*}{17.33} & \multirow{2}{*}{93.80} & \multirow{2}{*}{42.10} & \multirow{2}{*}{15.60} & \multirow{2}{*}{15.70} & \multirow{2}{*}{-3.25} & \multirow{2}{*}{-1.15} \\
         & Visual features & Grid & Global & & & & & & & & & \\
        \bottomrule
    \end{tabular}%
    }
    \caption{Ablation studies. "Overall $\Delta$" refers to the difference (ablated model - base model), averaged across datasets per task.}
    \label{tab:ablations}
\end{table*}

\subsection{Leveraging more shots}
\label{sec:more_shots}

In Table~\ref{tab:vqa}, we observe \Model's performance rapidly plateaus as the number of few-shot examples increases beyond 4. We hypothesize this could be related to the mapping network being trained on single image-caption pairs, and/or the visual embeddings still not being fully in-distribution with the language embeddings. Intuitively, a handful of examples may often help with task location~\cite{reynolds2021prompt}; however, the more shots are added, the more out-of-distribution the multimodal prompt becomes. This issue could be mitigated with in-context example selection~\cite{yang2021empirical} or better mixing of visual and textual modalities.

\subsection{0-shot VQAv2 results with OPT}
\label{sec:0shot_vqav2_opt}

In Table~\ref{tab:ablations2}, we observe \Frozen-OPT outperforms \Model-OPT on 0-shot VQAv2. Upon close inspection, we notice \Model-OPT often generates longer answers for yes/no questions, which receive a score of 0 according to VQA accuracy and VQAv2 reference answers -- this is a problem of the metric and not the model itself \cite{agrawal2022rethinking}. After filtering all answers starting with "yes" or "no" to leave only the short answer, \Model-OPT achieves a VQA accuracy of 40.14\% while \Frozen-OPT only reaches 32.03\%.

\subsection{4-shot results with OPT}

In Table~\ref{tab:ablations2}, we also observe that few-shot VQA performance is considerably lower for configurations using OPT-6.7B as language model. This is possibly due to the lack of a relative positional encoding~\cite{shaw2018self} in OPT, which is required for the transformer to generalize to prompt sequences where an image is not always in the first absolute position, or which contain more than one image \cite{tsimpoukelli2021multimodal}.

\subsection{Implementation details on simpler mapping networks}
\label{sec:simple_mappings_details}

In Sec.~\ref{sec:ablations}, we ablate the choice of mapping network architecture and replace it by simpler architectures. Similarly to \citet{eichenberg2021magma,merullo2022linearly}, the linear mapping is applied per-position on top of a flattened grid of visual features, and it projects from $D_i = 1024$ to $D_o = 4096$ dimensions (4.2M parameters). The output sequence length $L_o$ is thus equal to $L_i = 257$ (instead of $32$) -- as explained in Sec.~\ref{sec:architecture}, this increases the computational complexity in the subsequent LM, which in turn increases training and inference time considerably. Similarly to \citet{mokady2021clipcap}, the 2-layer MLP is applied on top of a global vector of visual features. The MLP's hidden dimensionality $D_h$ is equal to $D_i = 1024$, and the output dimensionality is $D_o = 32 * 4096$, which we split into $32$ vectors of $4096$ dimensions (135.3M parameters).

\subsection{Additional ablation studies}
\label{sec:more_ablations}

Table~\ref{tab:ablations} shows the results of our additional ablation studies. Unless specified otherwise, we perform all ablations on \Model$_\text{CC-clean}$ trained with 100\% of the data. These experiments are run only once and early stopping is based on the validation split of Conceptual Captions.

\paragraph{Pre-trained vision encoder.}
We ablate the pre-trained vision encoder used to compute image representations. We report results in row (i) of Table \ref{tab:ablations}. We compare two CLIP~\cite{radford2021learning} variants, our choice based on the ViT-L/14 backbone and the ViT-B/32 backbone. Indeed, the ViT-L/14 based vision encoder has an average +20\% advantage over the ViT-B/32 variant. We hypothesize this improvement is probably due to finer-grained image patches and a bigger model size.

\paragraph{Global vs. grid visual features.}
Grid features -- as opposed to global features -- preserve the spatial information in images. This kind of fine-grained information might be useful for complex VL tasks. To measure the impact of grid features, we train a version of \Model where we use the global image representation from CLIP's multimodal embedding space. Results are reported in row (ii) of Table \ref{tab:ablations}. We observe an average -10\% drop in performance, validating our choice of using grid over global visual features.

\paragraph{Mapping network architecture.}
We ablate the architectural design of our mapping network in rows (iii) and (v) of Table \ref{tab:ablations}. First, we ablate the size of our mapping network in terms of depth and hidden size. We explore three options: Small (2 layers and hidden size of 128), Medium (4 layers and hidden size of 256), and Large (8 layers and hidden size of 512). We see that using a smaller mapping network generally performs slightly worse than the base model. On the other hand, using a larger mapping network improves only in image captioning tasks, while increasing significantly the number of trainable parameters (from 3.4M to 19.5M). We also ablate the output sequence length $L_o$ of our mapping network. Similarly, reducing the output sequence length to 16 yields slightly lower performance overall, and increasing it to 64 only improves in image captioning tasks. In the extreme, we completely remove the learned constant embeddings and output the same sequence length coming from the vision encoder, i.e., $L_o = L_i = 257$. Following the trend, increasing the number of mapped visual embeddings is beneficial for image captioning but hurts VQA performance, while notably reducing training and inference throughput.

\paragraph{Data quality \& visual features.}
This ablation setting aims to be the most similar to Frozen: training on the \textit{full} (noisy) Conceptual Captions dataset while using global visual features. Results are reported in row (vi) of Table \ref{tab:ablations}. The overall performance is worse than that of our base model (-16\% on average), but still better than \Frozen$_\text{CC}$ on Table~\ref{tab:noisy_data} (+81\% on average). This validates our choice of using grid visual features while training on a subset of cleaner data.

\subsection{Additional qualitative results}
\label{sec:more_qualitative}

\begin{figure*}[ht]
    \centering
    \includegraphics[width=.99\linewidth]{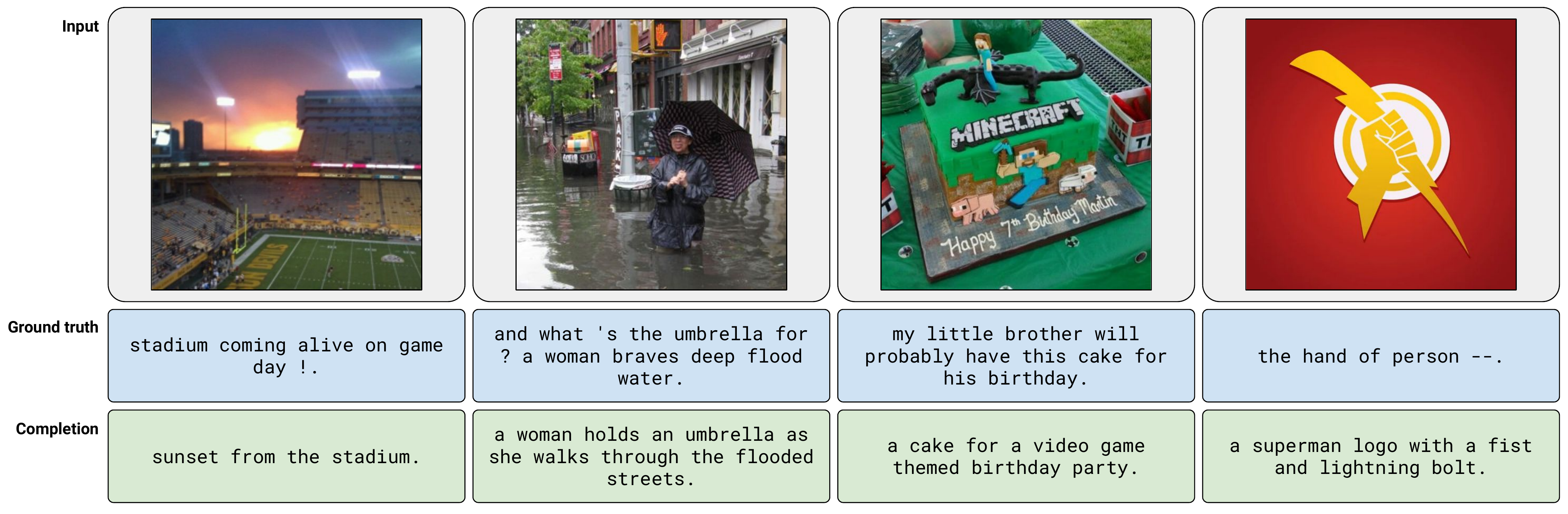}
    \caption{\Model's image captioning on Conceptual Captions.}
    \label{fig:cc_qualitative}
\end{figure*}

\begin{figure*}[ht]
    \centering
    \includegraphics[width=.99\linewidth]{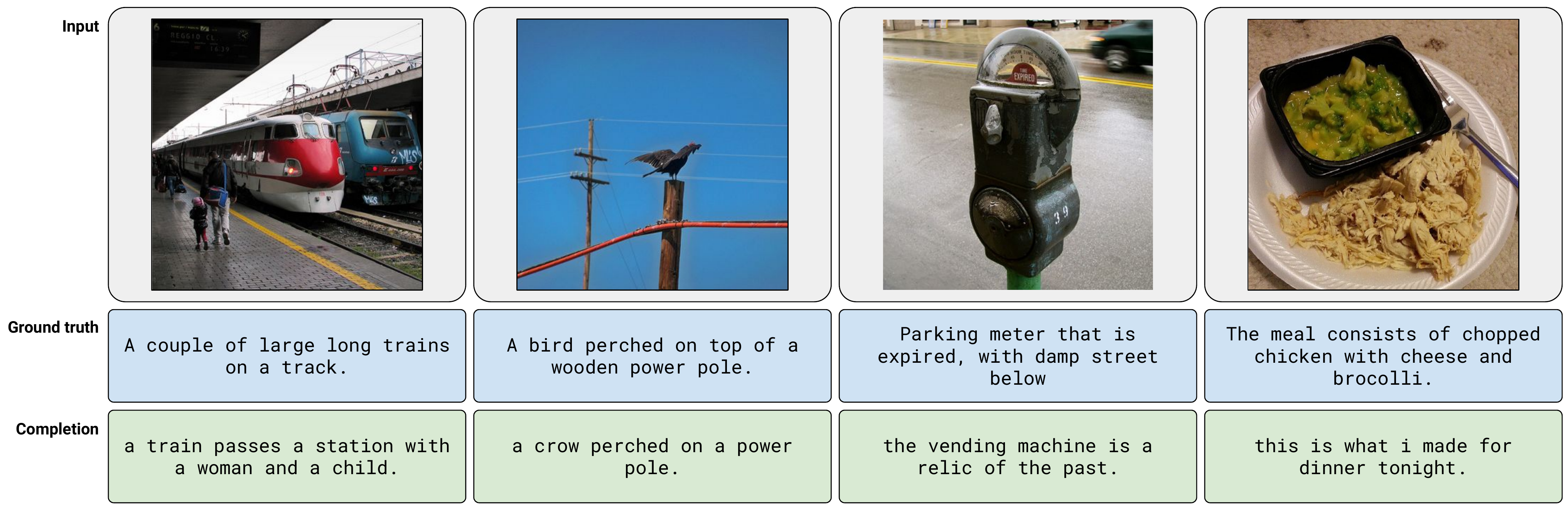}
    \caption{\Model's image captioning on COCO Captions.}
    \label{fig:coco_qualitative}
\end{figure*}

\begin{figure*}[ht]
    \centering
    \includegraphics[width=.99\linewidth]{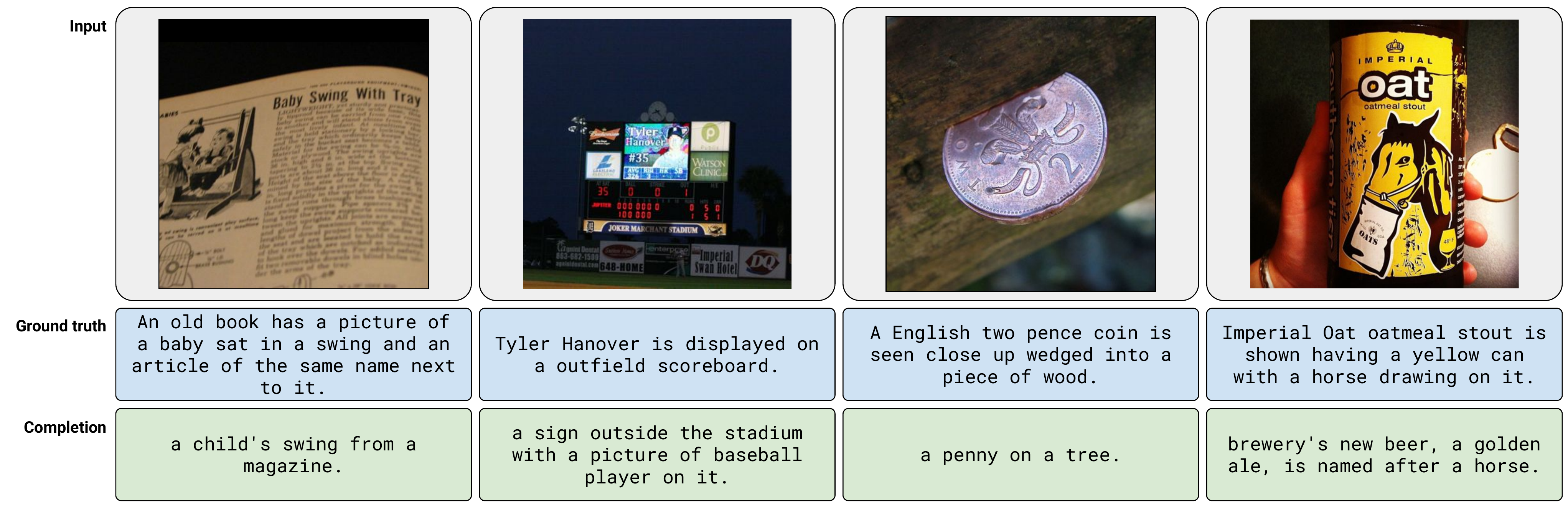}
    \caption{\Model's image captioning on TextCaps.}
    \label{fig:textcaps_qualitative}
\end{figure*}

\begin{figure*}[ht]
    \centering
    \includegraphics[width=.99\linewidth]{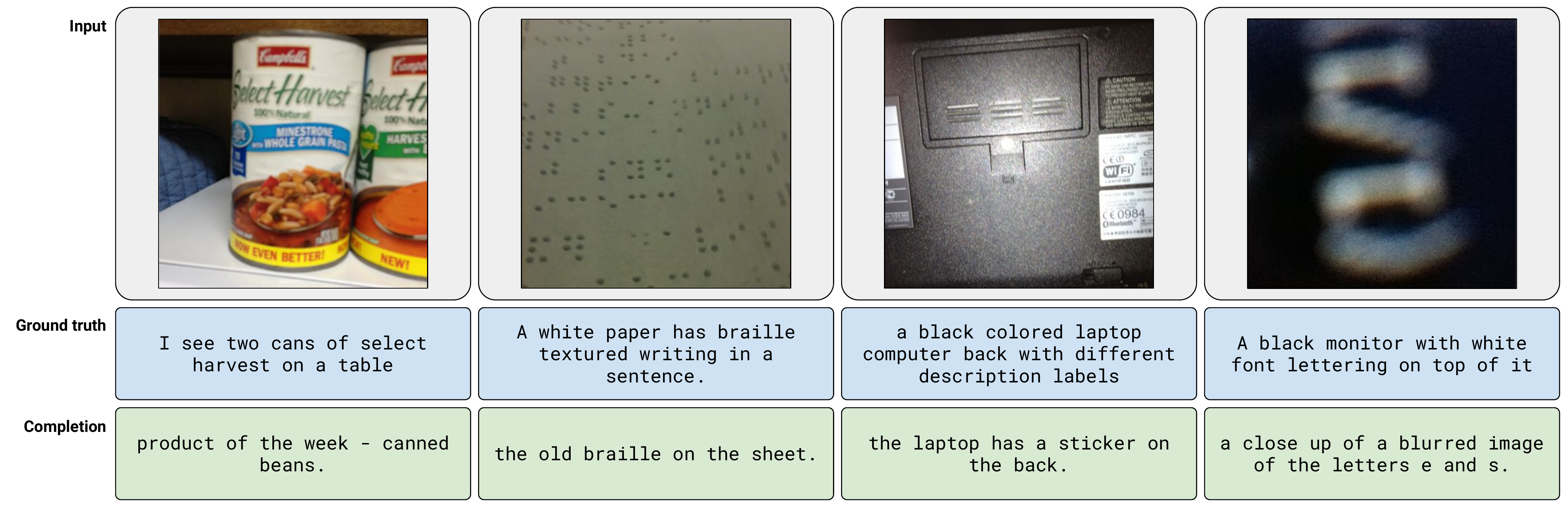}
    \caption{\Model's image captioning on VizWiz Captions.}
    \label{fig:vizwiz_qualitative}
\end{figure*}

\begin{figure*}[ht]
    \centering
    \includegraphics[width=.99\linewidth]{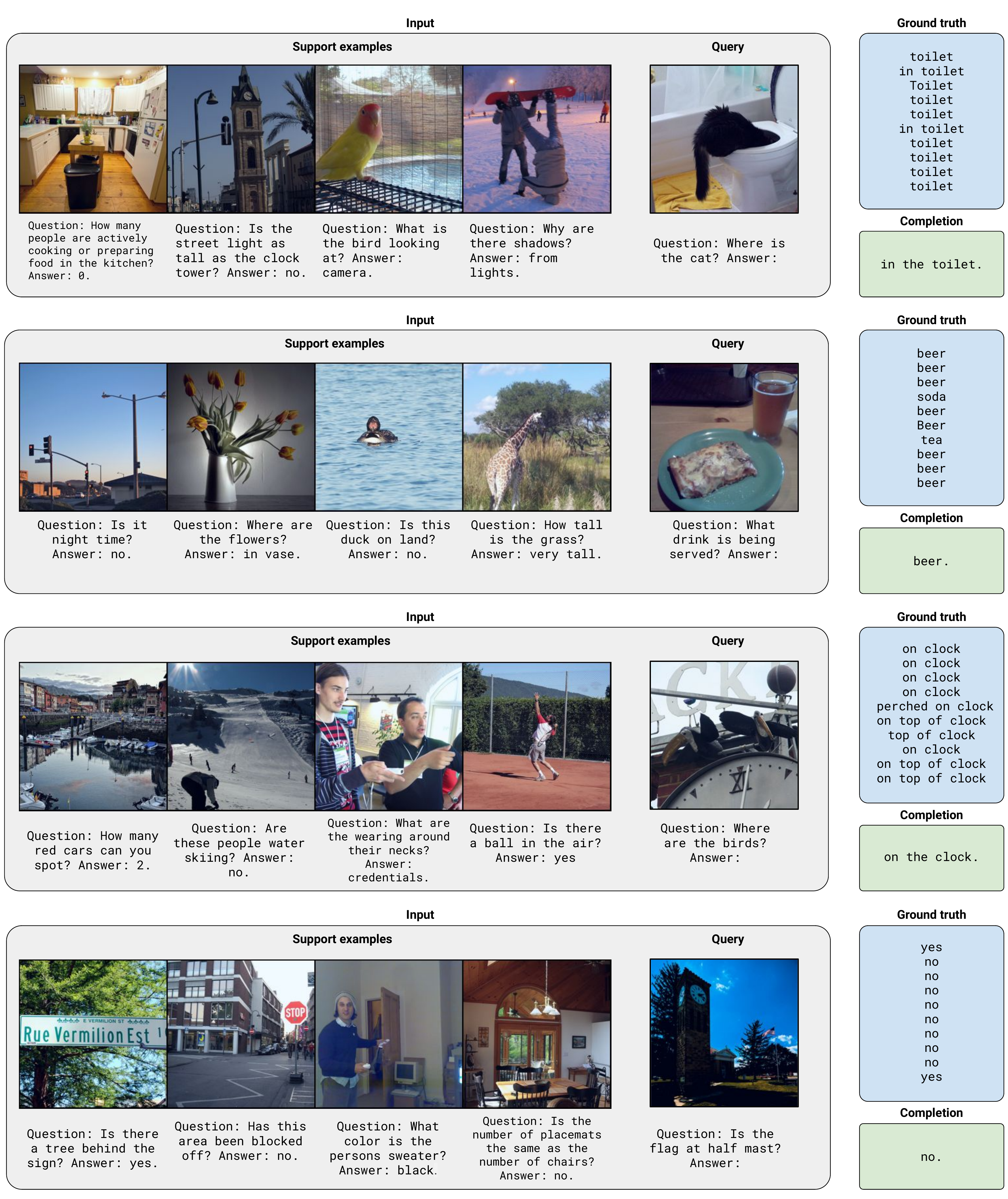}
    \caption{\Model's 4-shot VQA on VQAv2, success cases.}
    \label{fig:vqa_qualitative}
\end{figure*}

\begin{figure*}[ht]
    \centering
    \includegraphics[width=\linewidth]{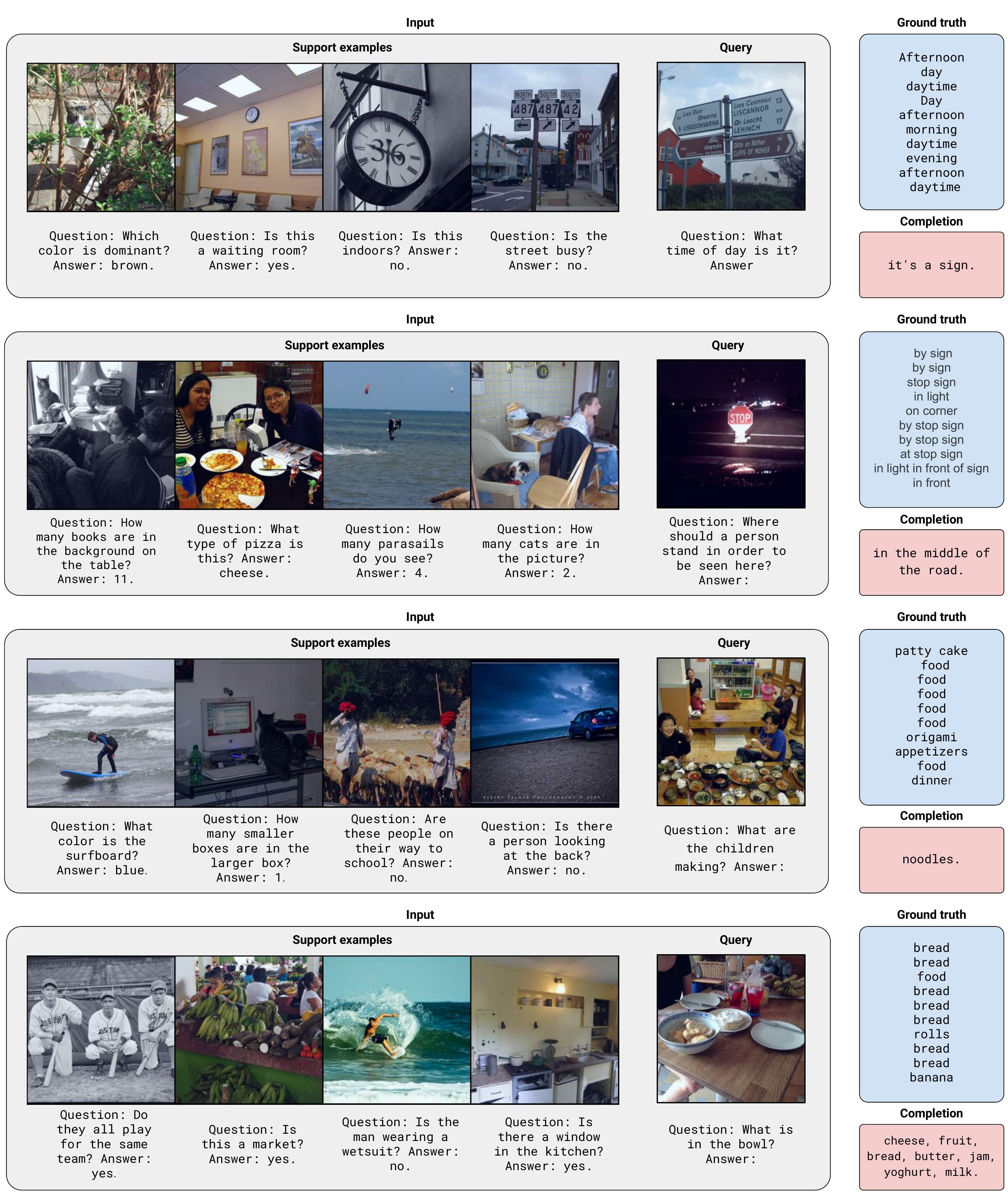}
    \caption{\Model's 4-shot VQA on VQAv2, failure cases.}
    \label{fig:vqa_failure_qualitative}
\end{figure*}

\begin{figure*}[ht]
    \centering
    \includegraphics[width=\linewidth]{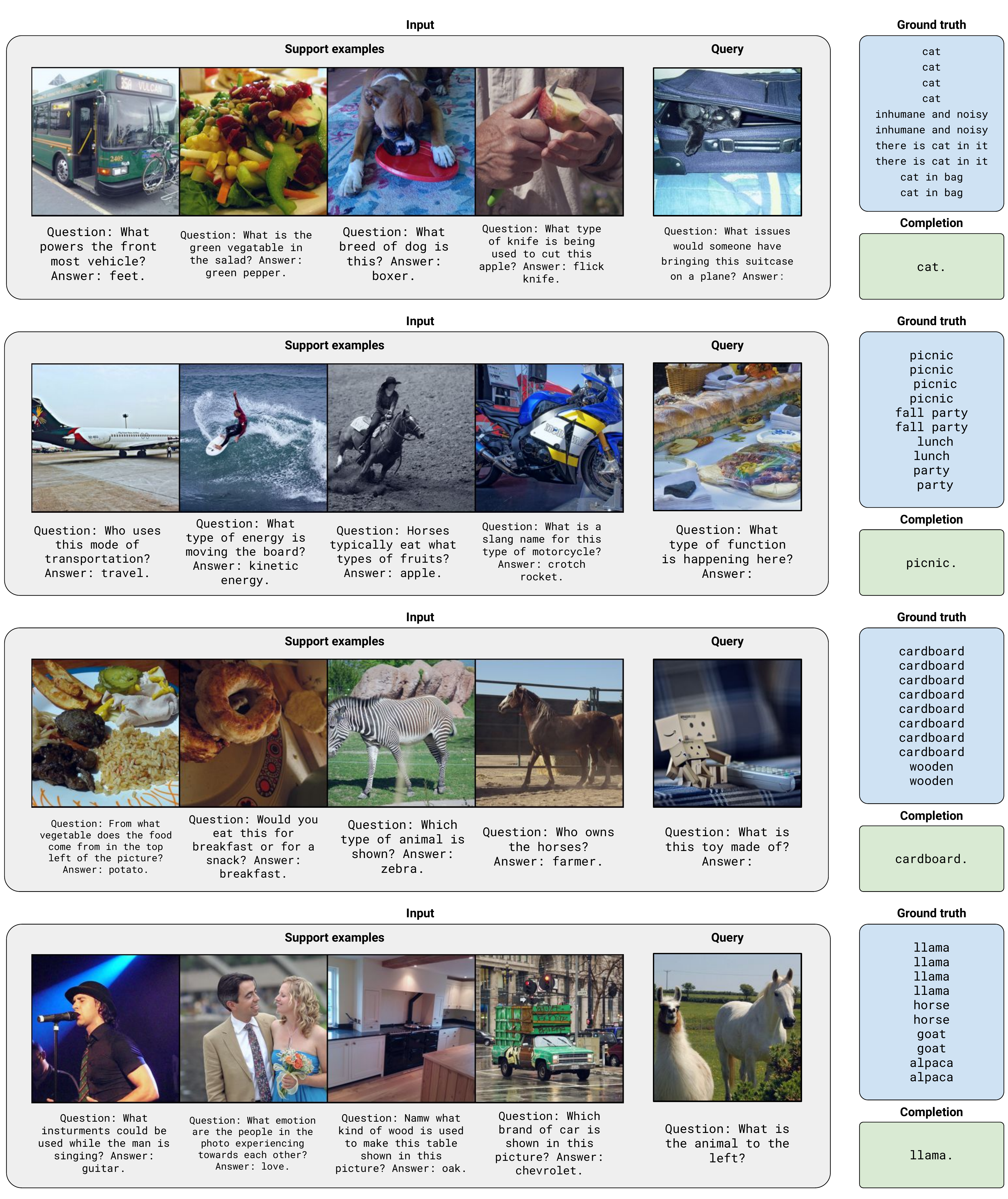}
    \caption{\Model's 4-shot VQA on OK-VQA, success cases.}
    \label{fig:okvqa_qualitative}
\end{figure*}

\begin{figure*}[ht]
    \centering
    \includegraphics[width=\linewidth]{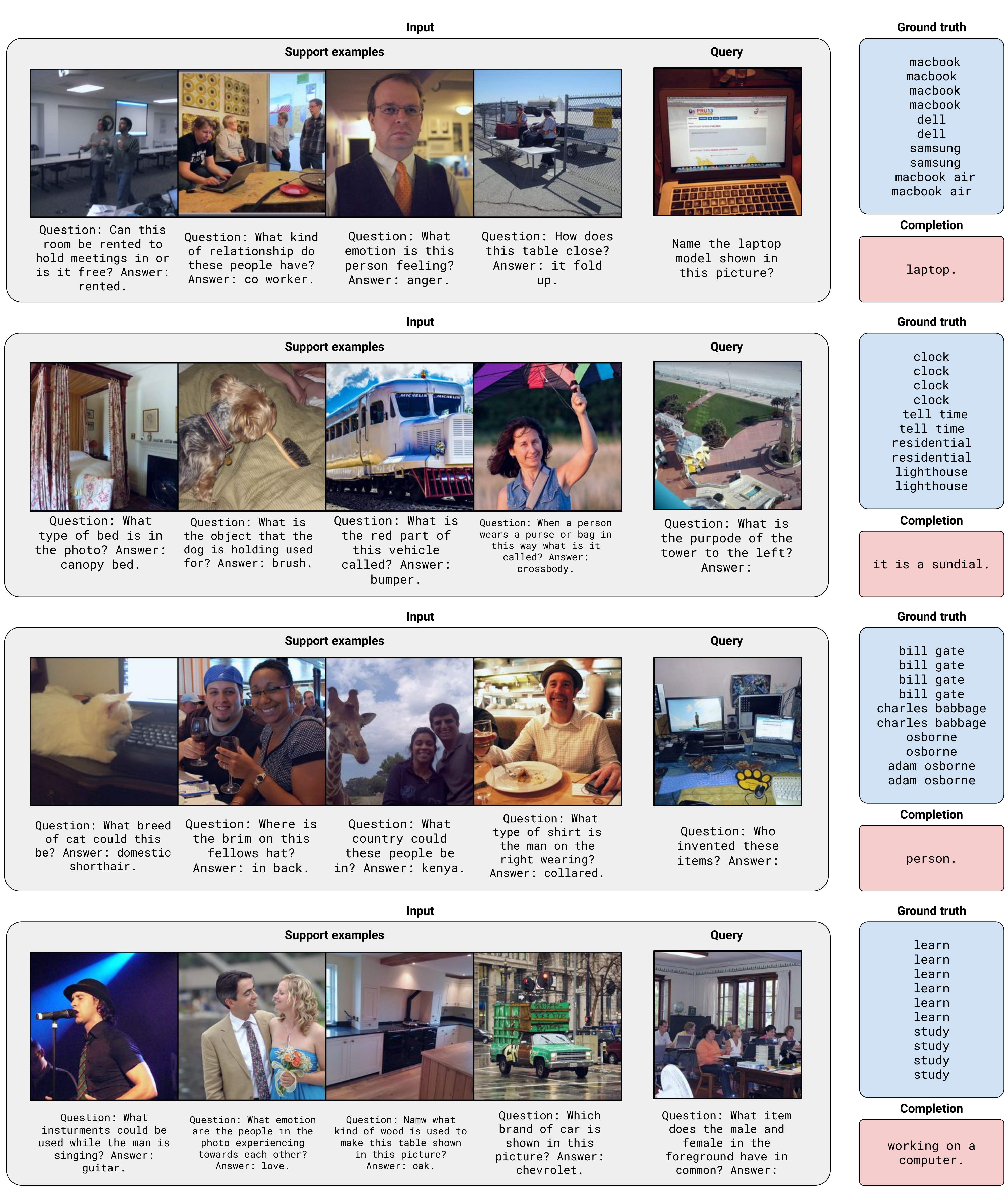}
    \caption{\Model's 4-shot VQA on OK-VQA, failure cases.}
    \label{fig:okvqa_failure_qualitative}
\end{figure*}

\begin{figure*}[ht]
    \centering
    \includegraphics[width=\linewidth]{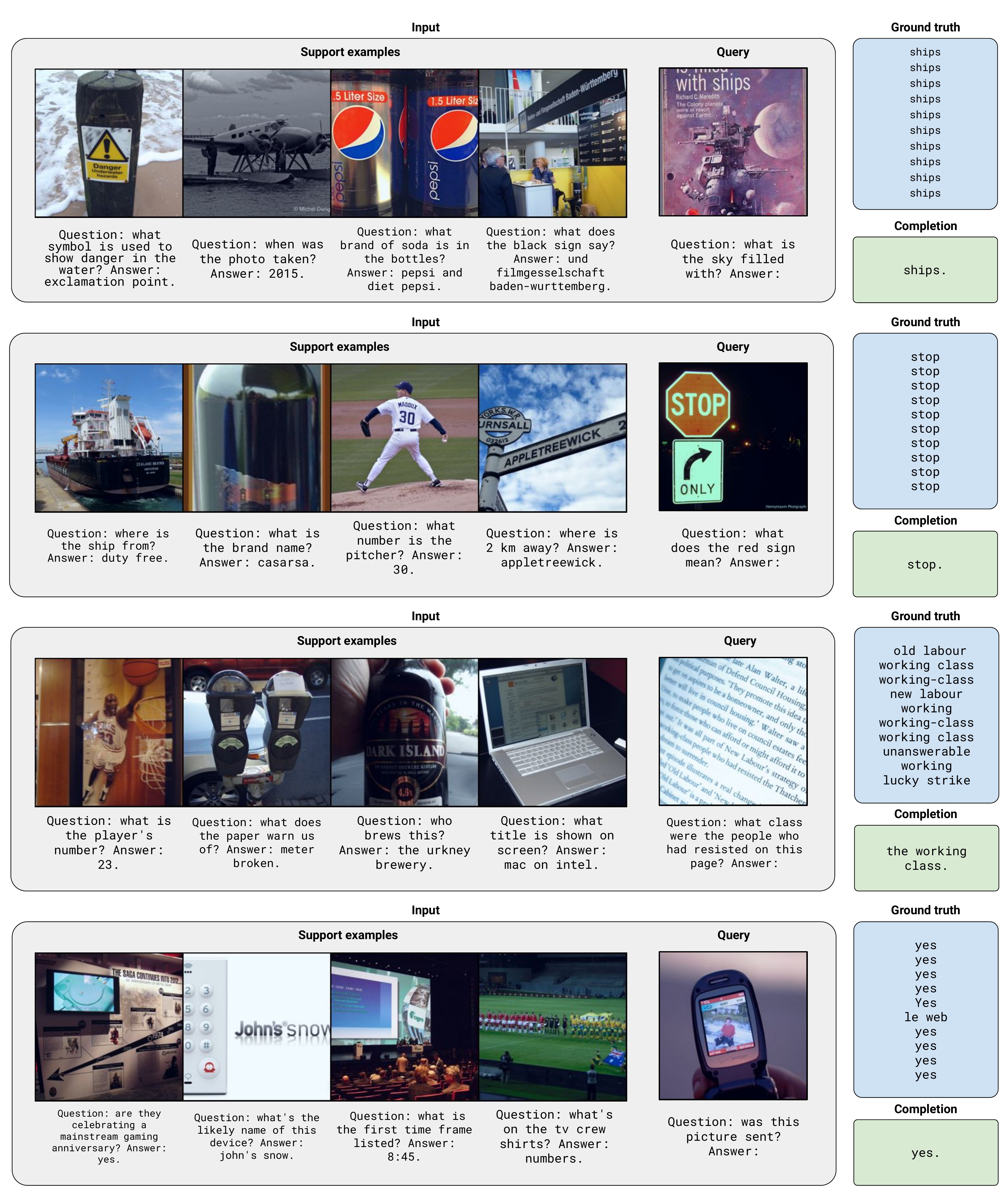}
    \caption{\Model's 4-shot VQA on TextVQA, success cases.}
    \label{fig:textvqa_qualitative}
\end{figure*}

\begin{figure*}[ht]
    \centering
    \includegraphics[width=\linewidth]{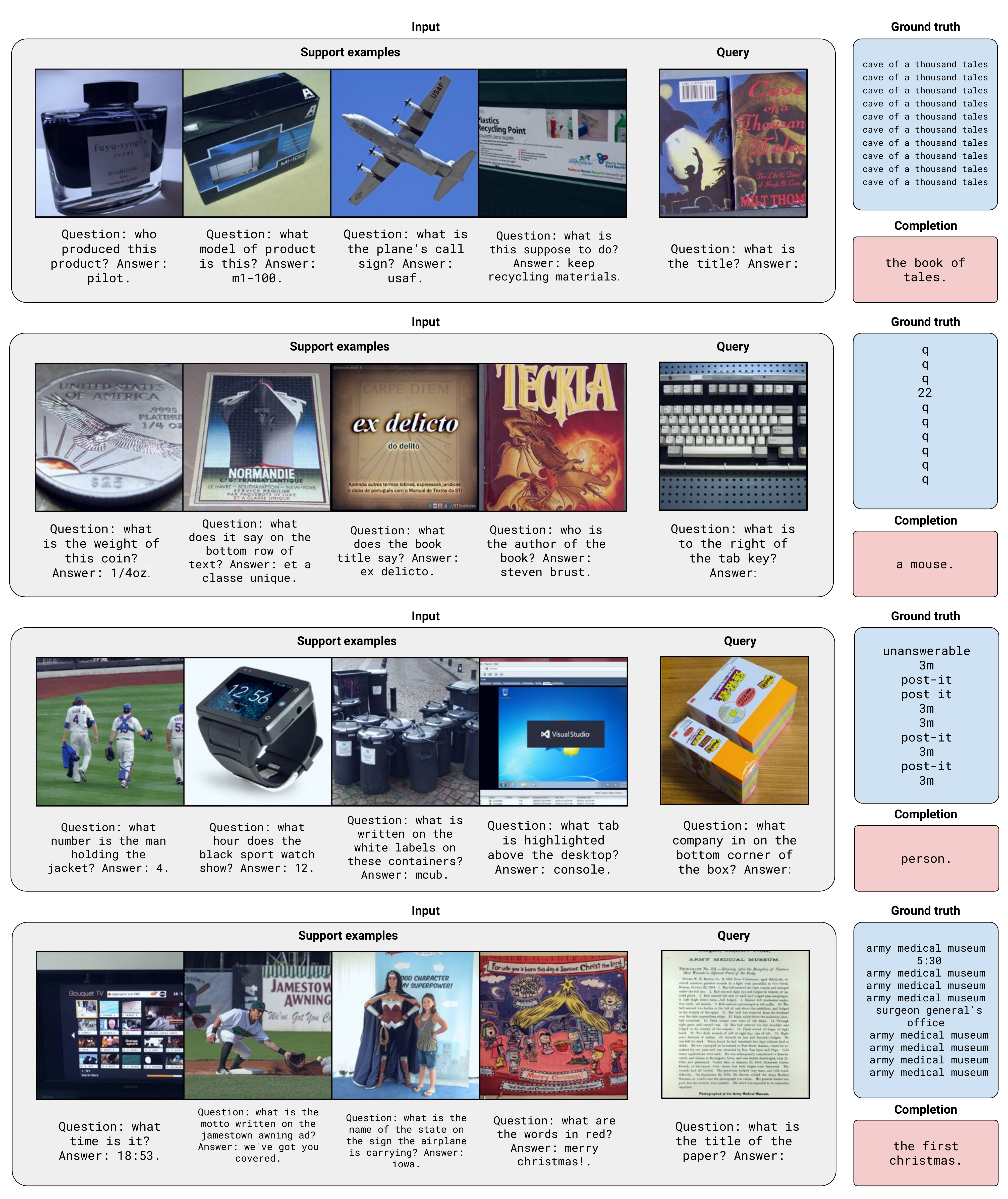}
    \caption{\Model's 4-shot VQA on TextVQA, failure cases.}
    \label{fig:textvqa_failure_qualitative}
\end{figure*}

\begin{figure*}[ht]
    \centering
    \includegraphics[width=\linewidth]{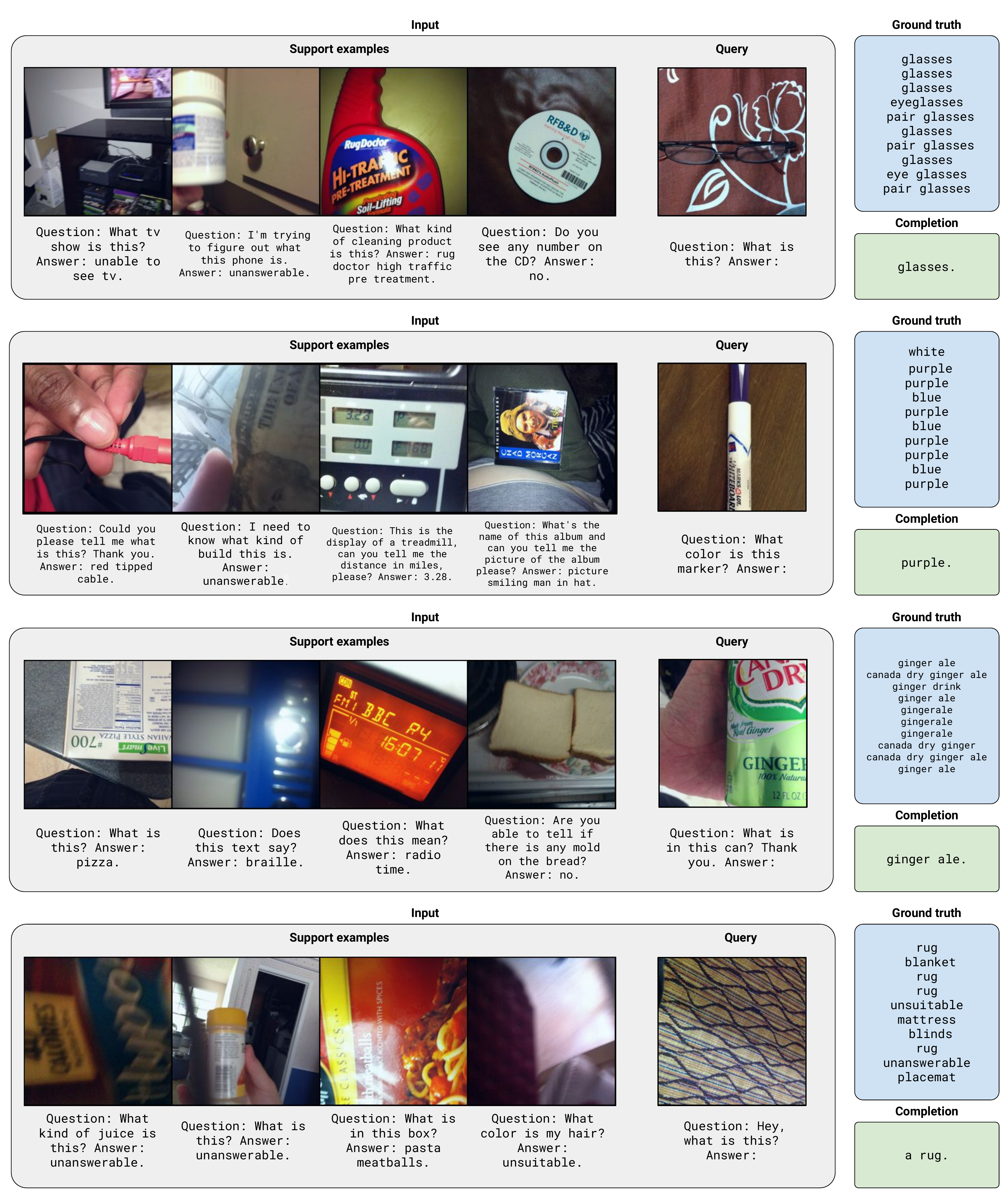}
    \caption{\Model's 4-shot VQA on VizWiz-VQA, success cases.}
    \label{fig:vizwizvqa_qualitative}
\end{figure*}

\begin{figure*}[ht]
    \centering
    \includegraphics[width=\linewidth]{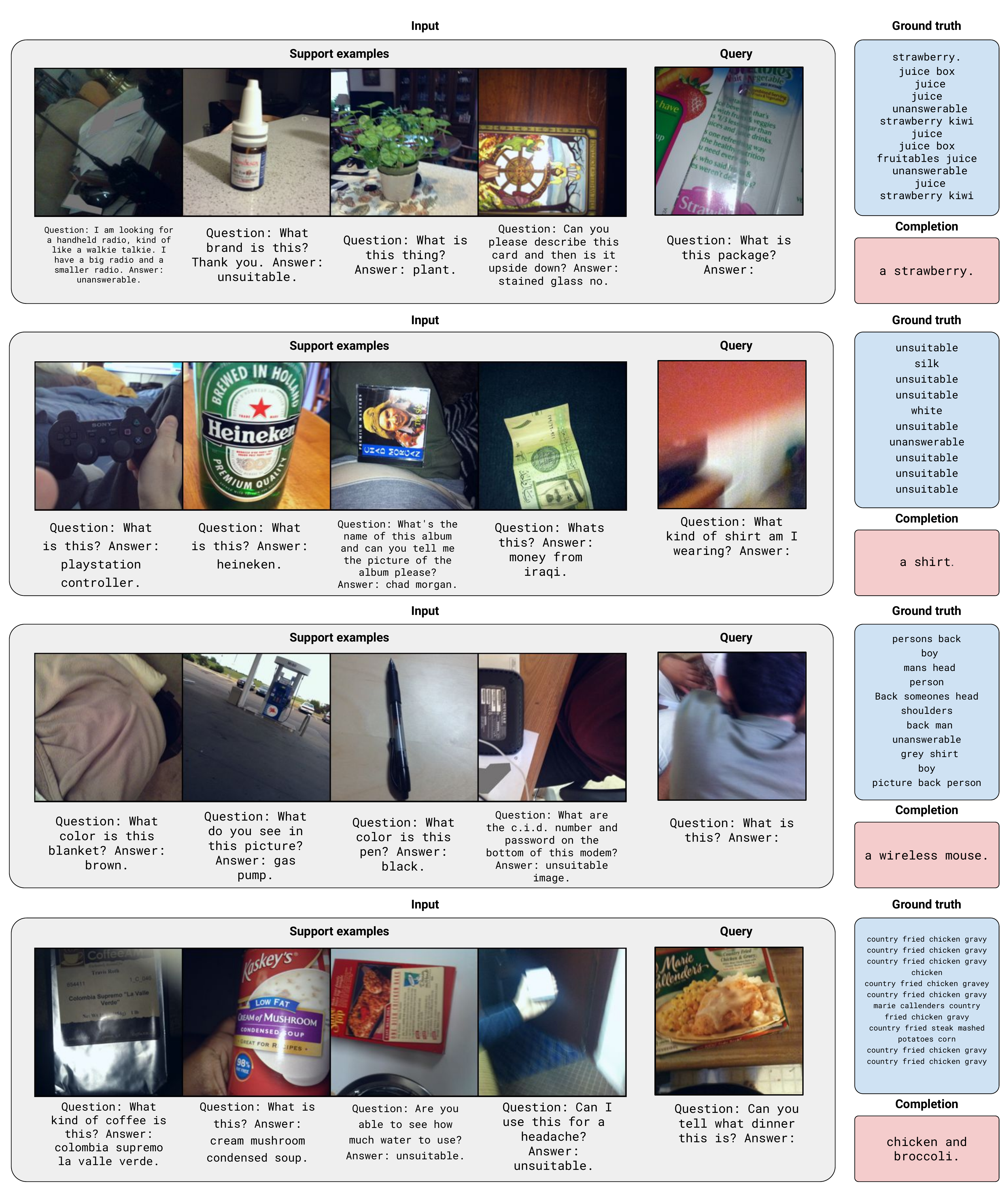}
    \caption{\Model's 4-shot VQA on VizWiz-VQA, failure cases.}
    \label{fig:vizwizvqa_failure_qualitative}
\end{figure*}

Figures~\ref{fig:cc_qualitative}-\ref{fig:vizwizvqa_failure_qualitative} show additional qualitative results of \Model$_\text{CC-clean}$ on random samples from different image captioning and VQA datasets. For VQA, in-context learning from 4 shots is performed.

\subsection{Interpretability of visual embeddings}

Using \Model$_{CC-clean}$, we extract mapped visual embeddings (after the mapping network) for ${\sim}30$ images from the COCO Karpathy-test set, and compute the nearest token embeddings (from the LM’s vocabulary) using cosine similarity. We rarely found the top-5 nearest tokens correspond to concepts present in the image, suggesting these embeddings are not interpretable. We hypothesize this is perhaps because they carry a combination of task-inducing and image-specific information, also pointed out by \citet{mokady2021clipcap}. We further cluster the mapped visual embeddings with K-means, and observe that each cluster often represents some visual concept (e.g., animals, food, sports). This means the mapped visual embeddings retain visual information from the vision encoder, which we also verify with \Model's performance on VL benchmarks.

\subsection{Analysis of VQA answer distributions}
\label{sec:qualitative_analysis}

\begin{figure*}[ht]
    \centering
    \includegraphics[width=\linewidth]{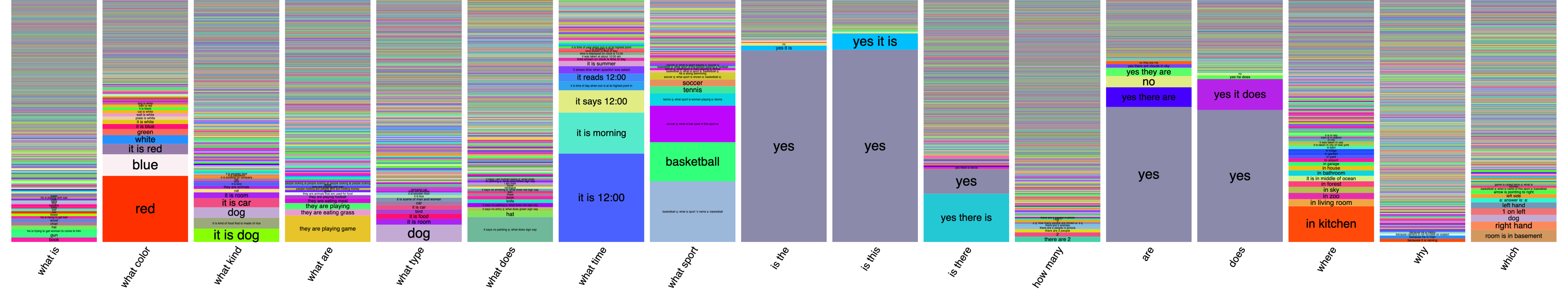}
    \caption{Predicted answer distributions for selected VQAv2 question types with the text-only baseline.}
    \label{fig:ans_dist_text_only}
\end{figure*}

\begin{figure*}[ht]
    \centering
    \includegraphics[width=\linewidth]{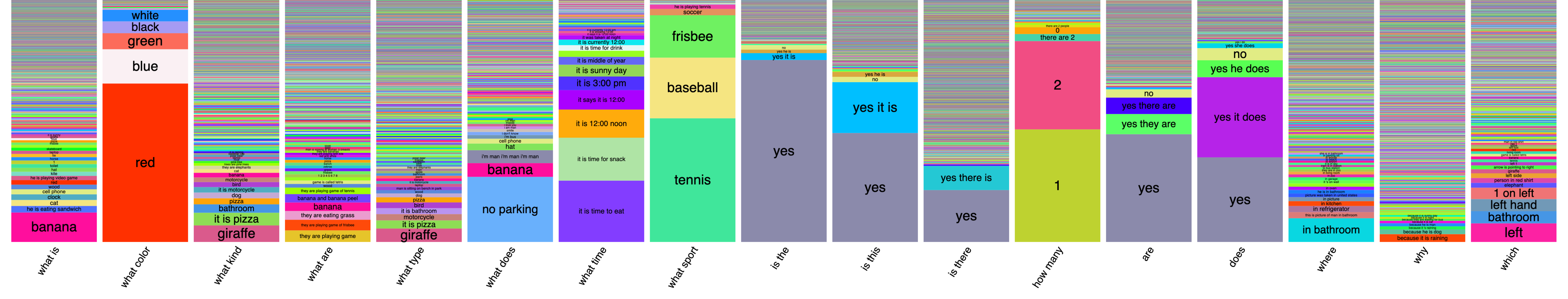}
    \caption{Predicted answer distributions for selected VQAv2 question types with the blind baseline.}
    \label{fig:ans_dist_blind}
\end{figure*}

\begin{figure*}[ht]
    \centering
    \includegraphics[width=\linewidth]{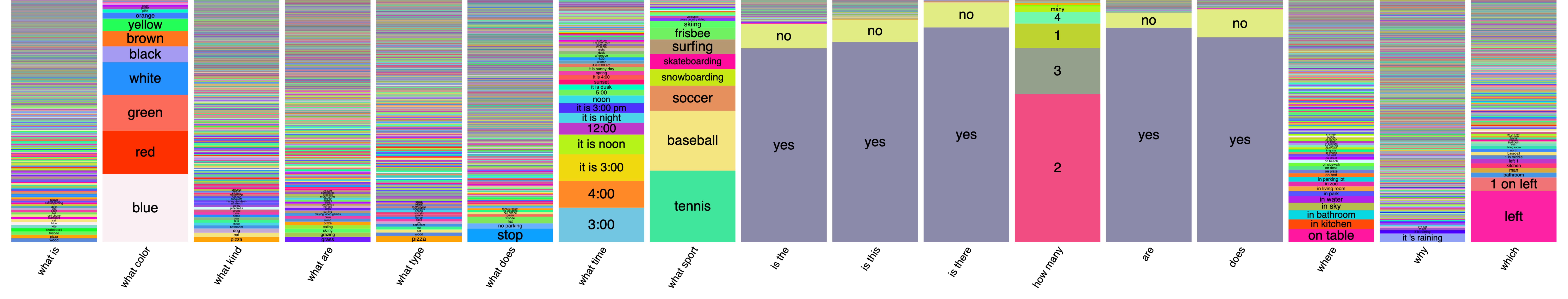}
    \caption{Predicted answer distributions for selected VQAv2 question types with \Model 4-shot.}
    \label{fig:ans_dist_mapl_4shot}
\end{figure*}

\begin{figure*}[ht]
    \centering
    \includegraphics[width=\linewidth]{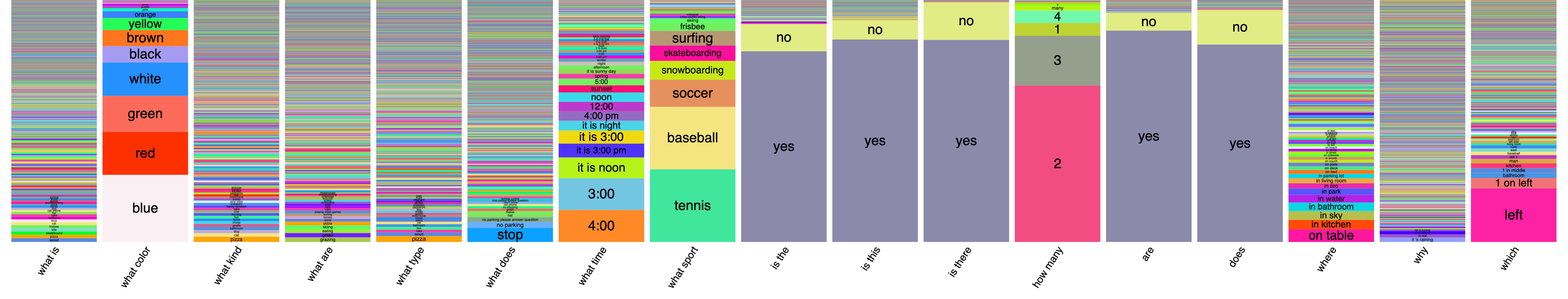}
    \caption{Predicted answer distributions for selected VQAv2 question types with \Model 8-shot.}
    \label{fig:ans_dist_mapl_8shot}
\end{figure*}

\begin{figure*}[ht]
    \centering
    \includegraphics[width=\linewidth]{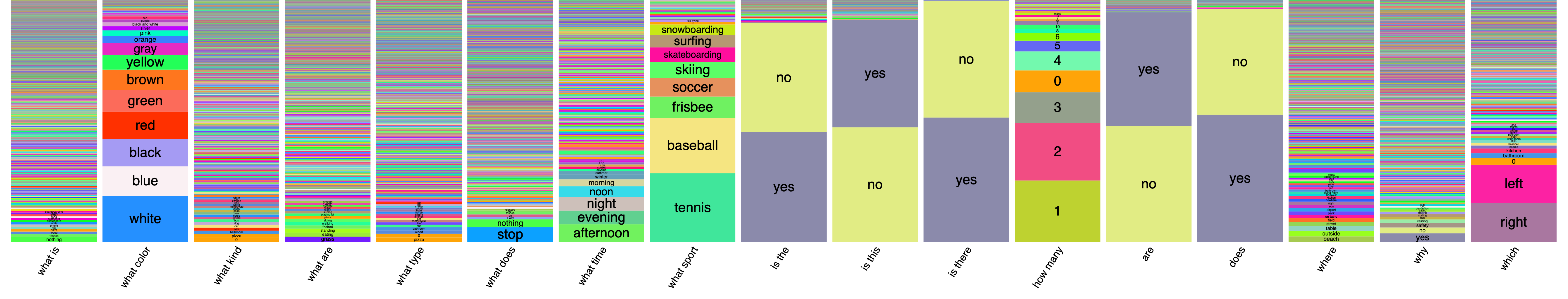}
    \caption{Ground truth answer distributions for selected VQAv2 question types.}
    \label{fig:ans_dist_ground_truth}
\end{figure*}

In this section, we show the distribution of answers for selected VQAv2 question types. We compare \Model with several baselines of our model to get insights into how the model's predictions change when training on increasing multimodal data. For the text-only baseline, we only provide the question text to the LM. This is different from the previously introduced blind baseline (Sec.~\ref{sec:experimental_setting}), where a blacked-out image is also provided. In particular, we compare the predicted answer distribution of \Model$_\text{COCO}$ evaluated on on zero- and few-shot VQA with the aforementioned baselines and the ground truth. Overall, we observe the predicted answer distribution gets closer to the ground truth answer distribution (Figure \ref{fig:ans_dist_ground_truth}) as more information from the image-question pair is provided to the model. We notice a considerable shift in the answer distribution from the text-only baseline (Figure~\ref{fig:ans_dist_text_only}) to the blind baseline, which demonstrates the impact caused by the captions alone. Moreover, we see the predicted answer distribution of \Model zero-shot is closer to the ground truth answer distribution than that of the blind baseline (Figure~\ref{fig:ans_dist_blind}), which indicates that \Model is leveraging the additional information from the visual input. For instance, we observe \Model's predicted answer distribution for the "what color" question type (second column) looks more similar to the ground truth distribution compared to the text-only and blind baselines. Finally, when performing in-context learning from four shots (Figure~\ref{fig:ans_dist_mapl_4shot}), we see the answer distribution gets even closer to the ground truth distribution. However, we do not observe much difference in answer distribution when increasing the number of shots from four to eight (Figure \ref{fig:ans_dist_mapl_8shot}).

\clearpage

\newpage

\newpage

\begin{figure*}[ht]
    \centering
    \includegraphics[width=\linewidth]{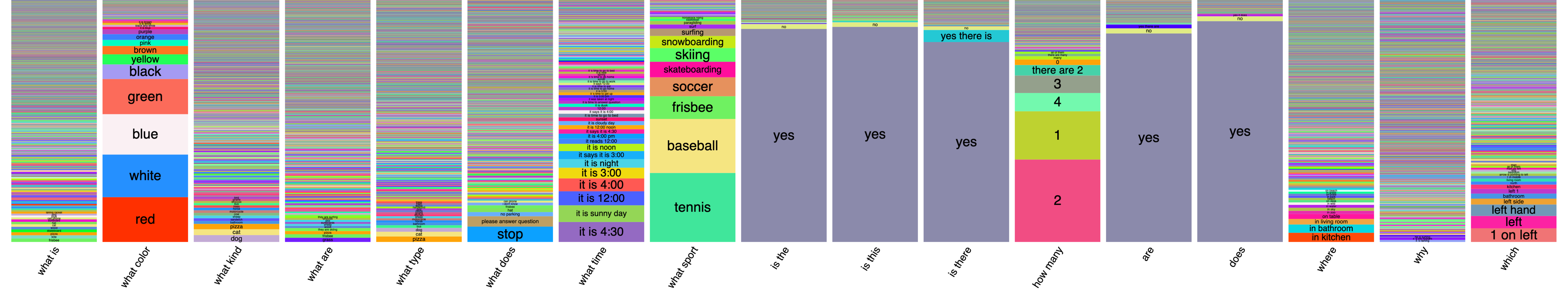}
    \caption{Predicted answer distributions for selected VQAv2 question types with \Model 0-shot.}
    \label{fig:ans_dist_mapl_0shot}
\end{figure*}

\end{document}